\documentclass[letterpaper]{article}
\usepackage[preprint]{paperstyle}
\usepackage{url}
\usepackage{graphicx}
\usepackage{natbib}
\usepackage{caption}
\usepackage{booktabs}
\usepackage{multirow}
\usepackage{xcolor}

\usepackage{amsmath}
\usepackage{amssymb}

\usepackage{enumitem}

\usepackage[breakable]{tcolorbox}
\tcbuselibrary{skins,breakable}

\usepackage{inconsolata}
\usepackage{listings}
\lstdefinestyle{toolbold}{moredelim=[is][\bfseries]{@}{@}}

\newtcolorbox{skillbox}[1]{%
  breakable,
  colback=gray!5,
  colframe=gray!60,
  arc=2mm,
  boxrule=0.8pt,
  fonttitle=\small\sffamily\bfseries,
  fontupper=\footnotesize,
  left=4pt,
  right=4pt,
  top=4pt,
  bottom=4pt,
  title={Domain: #1}
}

\newcommand{\skillfield}[1]{\textbf{\ttfamily #1:}}

\newtcolorbox[auto counter]{promptbox}{%
  enhanced jigsaw,
  breakable,
  colback=white,
  colframe=black!70,
  colbacktitle=black!10,
  coltitle=black,
  fonttitle=\bfseries,
  fontupper=\footnotesize,
  arc=6pt,
  title={Prompt~\thetcbcounter}
}

\newtcolorbox{judgebox}[1]{%
  enhanced jigsaw,
  breakable,
  colback=white,
  colframe=black!70,
  colbacktitle=black!10,
  coltitle=black,
  fonttitle=\bfseries,
  fontupper=\footnotesize,
  arc=6pt,
  title={#1}
}

\newtcolorbox{trajbox}[1][]{%
  enhanced jigsaw,
  breakable,
  colback=white,
  colframe=black!70,
  colbacktitle=black!10,
  coltitle=black,
  fonttitle=\bfseries,
  fontupper=\footnotesize,
  arc=6pt,
  #1
}

\newcommand{\trajphase}[1]{%
  \par\smallskip
  \noindent\rule{\linewidth}{0.4pt}\par\nobreak
  {\centering\footnotesize\bfseries\scshape #1\par}%
  \nobreak\noindent\rule{\linewidth}{0.4pt}%
  \smallskip
}

\newcommand{\turnlabel}[1]{\par\smallskip\noindent\textbf{#1}\par\nobreak\vspace{1pt}}

\title{Search2Skill: Skill Distillation Beyond Knowledge Boundaries \\Via Rubric-Based Reinforcement Learning}

\author{
    Muyang Ye\textsuperscript{\rm 1},
    Tian Lan\textsuperscript{\rm 2},
    Feihu Jiang\textsuperscript{\rm 2},
    Yongshi Ye\textsuperscript{\rm 3},
    Wuyunsiqin\textsuperscript{\rm 1},
    Bin Zhu\textsuperscript{\rm 2},
    Qianghuai Jia\textsuperscript{\rm 2},\\
    Zhao Xu\textsuperscript{\rm 2},
    Weihua Luo\textsuperscript{\rm 2},
    Ye Wang\textsuperscript{\rm 4},
    Jinyang Zhang\corresponding\textsuperscript{\rm 1},
    Longyue Wang\corresponding\textsuperscript{\rm 2},
    Lingfeng Bao\corresponding\textsuperscript{\rm 1}
}
\affiliations{
    \textsuperscript{\rm 1}Zhejiang University\quad
    \textsuperscript{\rm 2}Alibaba Group\\
    \textsuperscript{\rm 3}Institute of Artificial Intelligence, Xiamen University\quad
    \textsuperscript{\rm 4}Zhejiang Gongshang University
}

\def\correspondingtext{Corresponding authors. Contact: lingfengbao@zju.edu.cn}

\begin{document}

\maketitle

\begin{abstract}
Reusable skills, which encapsulate the procedural knowledge required to solve real-world
professional tasks, offer LLM-based agents a path toward self-evolution in expert domains. 
Existing self-evolving skill methods construct skills internally from the model's parametric
knowledge or trajectories, and are therefore bounded by what the model already knows. 
However, the domain conventions and standard procedures underlying professional skills often lie beyond
this boundary and are hard to elicit from the agent alone. 
To address this issue, we therefore propose a novel framework, Search2Skill,
that automatically identifies the agent's capability gaps, searches external sources to address them, and
distills the retrieved evidence into structured, reusable skills. 
Specifically, Search2Skill is optimized by a rubric-based reinforcement learning scheme that jointly improves 
when to search, how to search, and how to generate skills. 
Experiments on eight expert-level domains from three benchmarks show that Search2Skill 
consistently outperforms both search-augmented and trajectory-based skill-learning baselines 
under both streaming and held-out evaluation protocols. 
Further analyses show that the gains arise from skill abstraction rather than raw retrieved evidence, 
and that the acquired skills transfer across model scales.
\end{abstract}

\begin{links}
    \link{Code (to be released)}{https://github.com/ATH-MaaS/Marco-DeepResearch}
\end{links}

\section{Introduction}\label{sec:intro}
As LLM-based agents are deployed in increasingly open-ended and specialized environments, 
a central challenge is how they can continually expand what they are able to do~\cite{ni2026trace2skill,zhou2026comprehensive}. 
Reusable skills, procedural artifacts that encode domain-specific workflows, solution strategies, 
or tool-use knowledge \cite{skillzero2026}, offer agents a path toward self-evolution: 
once distilled, a skill can be reused to improve performance across an entire class of related tasks, 
a paradigm increasingly framed as \emph{self-evolving skills} \cite{yang2026skillopt,evoskill2026}.

However, existing skill learning remains largely \textbf{\emph{inward-looking}}: 
representative methods such as Trace2Skill distill skills purely by consolidating the agent's execution trajectories through inductive reasoning over its prior experience \cite{ni2026trace2skill,wang2023voyager,ouyang2025reasoningbank}, 
and thus cannot expand the model's underlying capability boundary. 
This is a fundamental limitation in professional domains, 
where the requisite skills, encoding domain conventions, standard procedures, and practical experience accumulated by human experts, often lie exactly beyond that boundary. 
Our preliminary study further reveals that models can often recognize when they lack the capability required for a task, 
yet this missing capability still causes a substantial drop in performance. 
We therefore argue that agent's self-evolving skill should also look \textbf{\emph{outward}}.
\begin{figure}[t]
    \centering
    \includegraphics[width=0.95\linewidth]{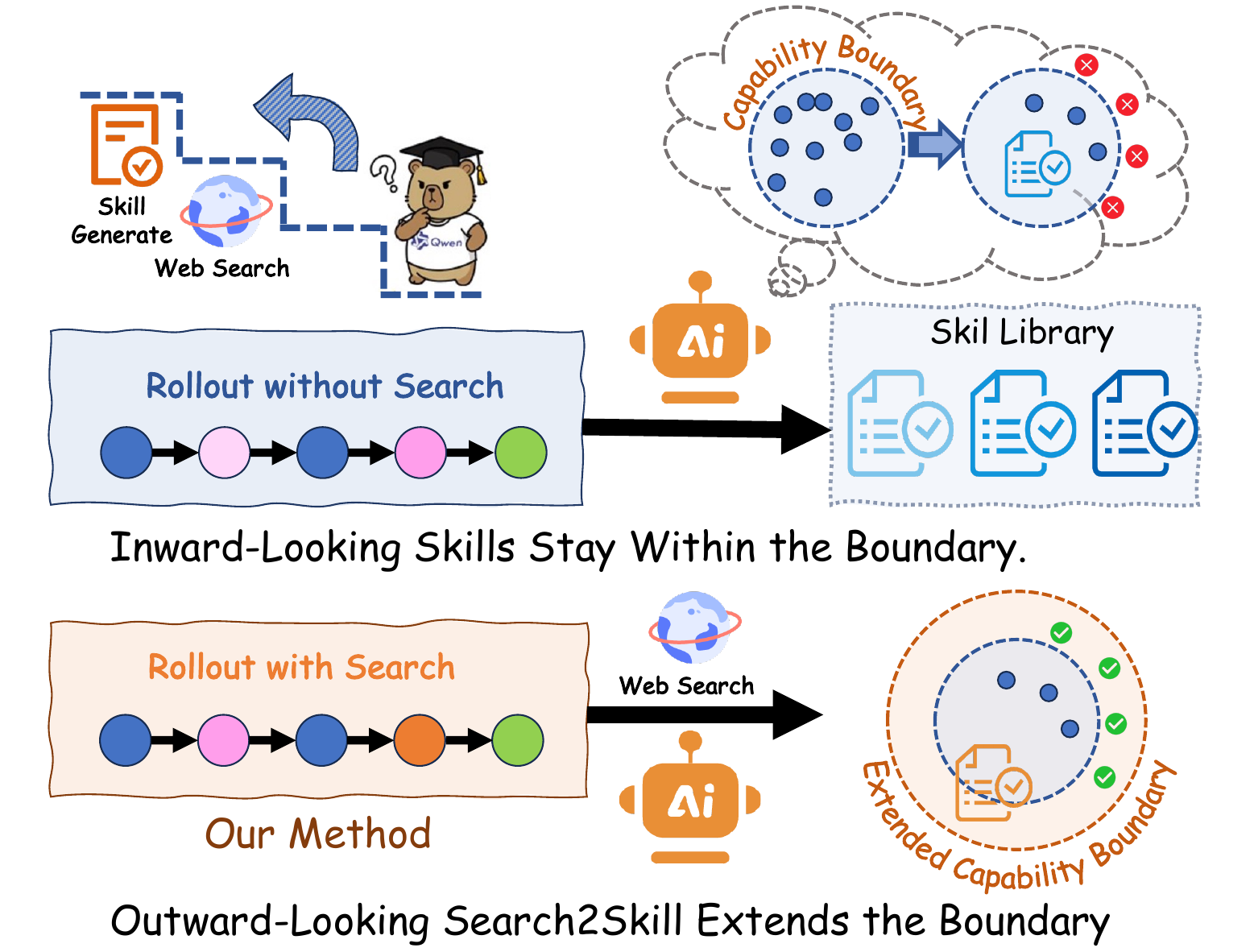}
    \caption{Inward-looking skill stays within the LLMs' boundary while outward-looking Search2Skill pushes it beyond.}
    \label{fig:teaser}
\end{figure}
As illustrated in Figure~\ref{fig:teaser}, a rollout without search can only recombine skills already inside the agent's capability boundary and therefore still fails once a task falls outside it, 
whereas a rollout with search lets the agent retrieve external evidence and distill what it finds into a new skill that pushes the boundary outward, turning the same previously unsolvable task into a solved one. 
Retrieving such knowledge from the web is often far more efficient than rediscovering it through trial-and-error experience, 
e.g., a niche physics identity that is readily available in the literature but rarely recovered through interaction alone.

To realize this idea, 
we propose \textbf{Search2Skill}, a search-driven skill acquisition framework that closes the loop among capability-gap identification, external search, and skill distillation. 
Given a task, the agent reasons with its parametric knowledge and skill library, and upon identifying a capability gap, 
enters an exploration state, formulates targeted queries, gathers evidence from external sources, and distills it into a skill that is written back to the library for both the current and future related tasks. 
Crucially, an effective agent must jointly learn \emph{when to search}, \emph{how to search}, and \emph{how to distill skills}; 
without optimization, an agent may misjudge when search is needed, retrieve irrelevant or biased evidence, or distill skills that are unfaithful to the evidence and overly instance-specific. 
We therefore optimize Search2Skill with a rubric-based reinforcement learning objective~\cite{gunjal2025rubrics,shao2026drtulureinforcementlearning}, beyond task success, 
rewards the exploration decision for calibrated necessity, queries for targeting the agent's capability gap, 
and skills for evidence grounding and generalizability. 
Unlike ordinary search-augmented answering~\cite{jin2025search,zhu2026marco}, which must re-search from scratch for every new task, 
the generalizable skill distilled by Search2Skill persists in the library and can be directly reused on future related tasks, 
continuing to expand the agent's effective capability boundary.

Extensive experiments on eight expert-level domains from three benchmarks, SuperGPQA~\cite{du2026supergpqa}, MMLU-Pro~\cite{wang2024mmlu}, and EvoAgentBench~\cite{gao2026evoagentbench}, 
demonstrate that Search2Skill consistently outperforms both search-augmented and trajectory-based skill-learning baselines. 
Under the \emph{streaming} evaluation protocol, Search2Skill improves average accuracy over direct inference 
by 8.3\% and 9.3\% on Qwen3-4B and Qwen3-8B, 
and performance improves by 5.1\% and 6.6\% under the stricter \emph{held-out} evaluation protocol. 
Further analyses show that skill abstraction itself drives these gains: 
reusing abstracted skills outperforms caching raw retrieved evidence by 4.5\%. 
Besides, the acquired skills could transfer across model scales: 
a skill library mined by the 8B model improves both smaller 4B and larger 14B executors, 
surpassing all strong baselines.

\section{Related Work}\label{sec:related_work}

\paragraph{Skill learning from experience.}
Self-evolving skill or memory equips LLM agents with reusable experience external to its frozen parameters,
letting it keep improving at test time \cite{zhou2026comprehensive}.
Existing methods realize this purely by distilling an agent's own trajectories.
Most commonly, this is done by prompting an off-the-shelf LLM to consolidate trajectories into skills or memory,
e.g., executable skill libraries \cite{wang2023voyager,zheng2024synapse,ni2026trace2skill,evoskill2026},
and reusable memory entries \cite{zhao2024expel,ouyang2025reasoningbank,fang2026memp}.

\paragraph{Reinforcement learning for self-evolving memory or skill.}
Because such prompting-based distillation tends to yield instance-specific experience, a growing line instead trains a dedicated model with reinforcement learning to curate skills or memory more reliably.
For example, UMEM trains a single model to jointly extract and manage generalizable memory using semantic neighborhood modeling~\cite{ye2026umem},
alongside similar systems that jointly optimize skill selection, usage, and distillation \cite{xia2026skillrl,tu2026dynamic,shi2026skill1,ouyang2026skillos,skillzero2026, huang2026skillselfplaypushingfrontier}. 
EvolveR~\cite{wu2026evolverselfevolvingllmagents} distills its own trajectories into experience to better perform deep search tasks.
However, both lines of methods above only distill experience from an agent's own trajectories and therefore cannot break through its capability boundary.
In contrast, Search2Skill's joint objective spans deciding when and how to search external knowledge,
capturing expert practical experience that such self-contained curation cannot recover.

\paragraph{External knowledge augmentation.}
The most common approach for extending an agent's capability boundary is search-augmented agents that search and verify evidence at inference time~\cite{jin2025search,zhu2026marco,chen2026agentcpmexplore,tongyi2025deepresearch}.
However, such agents only solve current queries, and the retrieved evidence is instance-specific.
Compared with these methods, Search2Skill explicitly strengthens the agent's ability to distill generalizable skills from search results.
Experimental results in Analysis Section confirm that directly reusing raw evidence is too instance-specific to generalize, while abstracted skills distilled by Search2Skill bring a 4.5\% average gain over raw evidence (Table~\ref{tab:web_vs_skill}).

\section{Preliminary Study}\label{sec:preliminary}

LLMs are largely aware of their own capability boundary, though not perfectly~\cite{kadavath2022language,yin2023large}.
We study this in the context of search-driven skill acquisition, on a subset of SuperGPQA-Science annotated with teacher reference trajectories, and obtain two findings.

\begin{figure}[t]
    \centering
    \includegraphics[width=\linewidth]{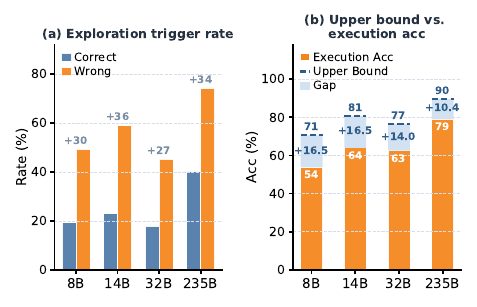}
    \caption{Knowledge-boundary awareness across Qwen3 model scales on SuperGPQA-Science:
(a) Exploration trigger rate on eventually correct and incorrect questions;
(b) Execution accuracy and upper bound of self-evolving.
}
    \label{fig:preliminary}
\end{figure}

\paragraph{Models (mostly) anticipate their own failures.}
When solving a task, the agent can either answer directly or enter an exploration state to search externally for missing knowledge.
We compare how often exploration is triggered on questions the agent eventually answers correctly versus incorrectly.
As shown in Figure~\ref{fig:preliminary}(a), across four Qwen3 models (8B/14B/32B/235B) the trigger rate on incorrect questions exceeds that on correct ones by 27 to 36 points,
confirming that models carry a meaningful, if imperfect, signal about their own capability boundary before attempting a question.
Directing external search specifically at these recognized weak spots could make self-evolution far more efficient.

\paragraph{Closing this gap would raise the ceiling of self-evolution.}
We next ask how much headroom this signal leaves on the table.
For each exploration-triggered, incorrectly answered question, 
the agent states which capability it believes is missing, and a teacher model Qwen3.7-Max~\cite{qwen37blog} independently diagnoses the true missing capability from the agent's multiple trajectories and the ground-truth solution.
We count a question toward the upper bound only when the two diagnoses match, i.e., the agent is not just stuck but correctly knows why; its realized accuracy defines the \emph{execution} boundary.
Figure~\ref{fig:preliminary}(b) shows this gap averages $14.3\%$ across Qwen3 models: closing it would raise 14B and 32B to 81\% and 77\%, approaching or even exceeding the 235B model's 79\% execution accuracy.
While existing self-evolving approaches cannot close this gap, since they only reorganize experience already inside the agent's boundary, 
targeting these recognized gaps with external search and distilling the resulting experience into reusable skills can raise both the efficiency and the ceiling of self-evolution, 
which is exactly the goal of the Search2Skill framework introduced next.

\begin{figure*}
    \centering
    \includegraphics[width=\linewidth]{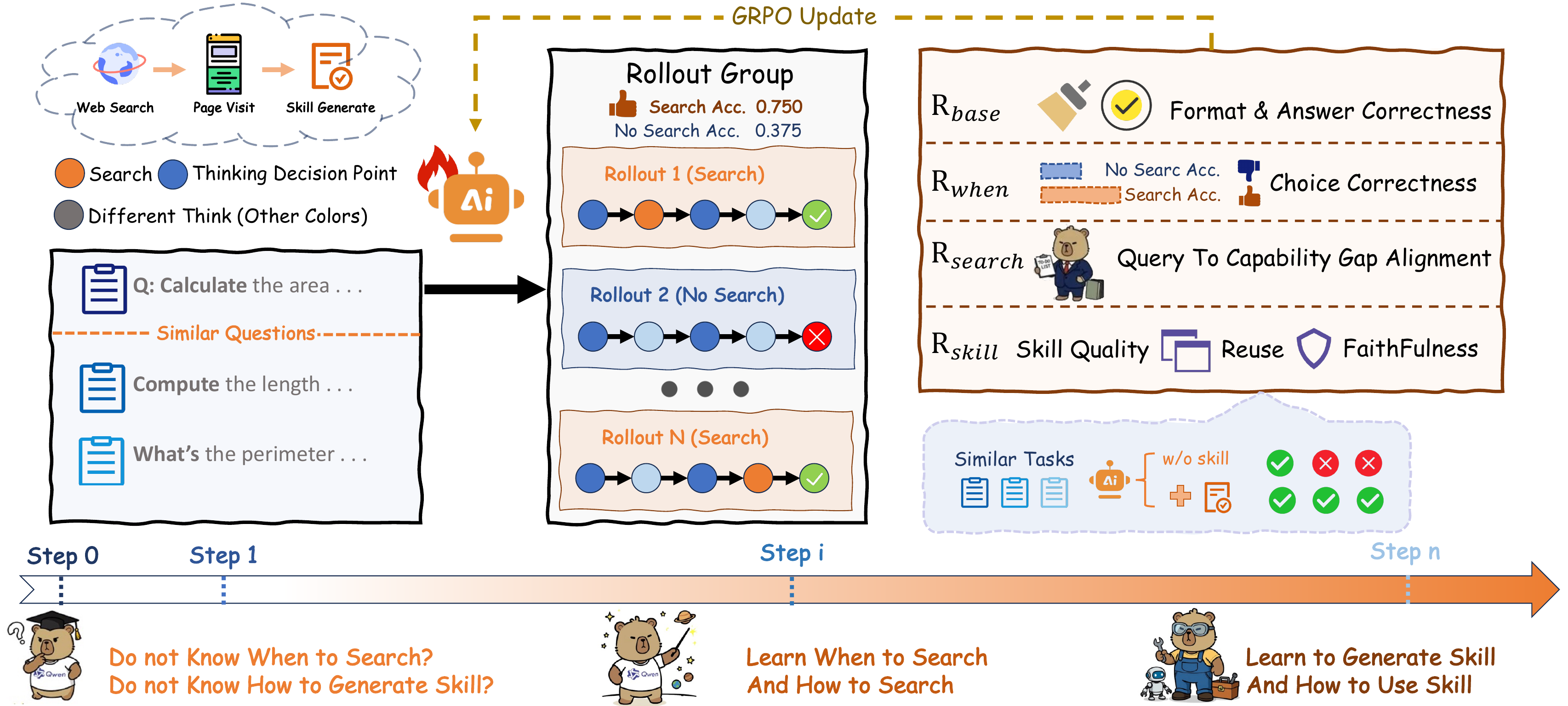}
    \caption{Overview of Search2Skill and its reward design. The agent searches when needed, distills retrieved evidence into reusable skills, and is trained with rewards for \emph{when to search}, \emph{how to search}, and \emph{how to generate skills}.}

    \label{fig:framework}
\end{figure*}

\section{Methodology of Search2Skill}\label{sec:method}
\subsection{Task Formulation of Search2Skill Framework}\label{sec:method:framework}

Search2Skill augments a language agent with \emph{external exploration} and a persistent \emph{skill library}, enabling the agent to convert retrieved evidence to reusable procedural knowledge. 

Formally, we formalize self-evolving skill acquisition as a task defined by the tuple $(q, \mathcal{M})$, where $q$ is the input question and $\mathcal{M}$ is a persistent skill library accumulated from the agent's past experience.
Within this task, Search2Skill instantiates the agent's policy.
Relevant skills $\mathcal{M}_q \subseteq \mathcal{M}$ are automatically retrieved and injected into the prompt context prior to execution.
Conditioned on $q$ and $\mathcal{M}_q$, the agent conducts an interaction trajectory
\begin{equation}
\tau = (o_1, r_1, a_1, o_2, r_2, a_2, \dots, o_T, r_T, a_T),
\end{equation}
where $o_t$ is the observation returned by the environment, $r_t$ is the agent's intermediate reasoning, and $a_t$ is the action taken at step $t$.
The action space covers three decisions central to Search2Skill:
(1) a \emph{search} action, i.e., a web search or browsing operation (e.g., \texttt{search} or \texttt{visit}) that queries external web resources when the agent identifies a capability gap~\cite{zhu2026marco};
(2) a \emph{skill-generation} action that distills the retrieved evidence into a structured skill $s$ and updates the library, $\mathcal{M} \leftarrow \mathcal{M} \cup \{s\}$;
and (3) a \emph{final-answer} action that terminates the rollout.
At each step, the policy $\pi$ generates the next reasoning-action pair conditioned on the question, the trajectory so far, and the skill library,
\begin{equation}
r_{t+1}, a_{t+1} \sim \pi(\cdot \mid q, \tau_t, \mathcal{M}).
\end{equation}

If the agent does not explore, the trajectory reduces to a standard reasoning-and-answering rollout. 
If it explores, the trajectory additionally includes search and skill-generation actions to distill the retrieved evidence into a structured skill $s$. 
After the rollout terminates, the generated skill is written back to the skill library and can be reused both for the current question and future related tasks.

\subsection{Reward Design for Search2Skill Optimization}\label{sec:method:rl}

Search2Skill's policy must decide \emph{when to search}, \emph{how to search}, and \emph{how to distill a skill}, and each of these three decisions must be carefully optimized.
Mainstream GRPO-based agent training, however, relies on a single outcome-based reward tied to final task success~\cite{jin2025search,ye2026umem}, 
lacking fine-grained supervision over these three decisions and thereby inducing a credit-assignment problem.
To address this, we design a rubric-based RL objective~\cite{gunjal2025rubrics,shao2026drtulureinforcementlearning} that supervises each of these three decisions with a dedicated reward term:
$R_{\text{when}}$ decides whether exploration is necessary,
$R_{\text{search}}$ evaluates the quality of the search queries,
and $R_{\text{skill}}$ evaluates the generalizability and faithfulness of the generated skill,
on top of a task term $R_{\text{base}}$ that checks final-answer correctness and action-format correctness (Figure~\ref{fig:framework}).
We detail each term below, and then compose them into the final rollout reward.

\paragraph{Exploration-Necessity Reward ($R_{\text{when}}$).}
To discourage both unnecessary exploration and over-confident direct answering, we judge exploration necessity \emph{relatively} within each rollout group rather than with a fixed bonus.
For each question $q$, we partition the sampled rollouts into an exploring subset $G_{\text{exp}}$ (exploring at least once) and a direct-answering subset $G_{\text{direct}}$, with mean accuracies $\bar{y}_{\text{exp}}$ and $\bar{y}_{\text{direct}}$ and gap $\Delta = \bar{y}_{\text{exp}} - \bar{y}_{\text{direct}}$.
Given a margin $\tau_0 > 0$, if exploration wins ($\Delta > \tau_0$), every exploring rollout becomes eligible for the quality bonus $q(\tau)$ defined below;
if direct answering clearly wins ($\Delta < -\tau_0$), every direct rollout receives a unit bonus;
otherwise no decision bonus is granted, including the case where the group collapses to a single strategy.

\paragraph{Search-Quality Reward ($R_{\text{search}}$).}
We employ a rubric-based LLM-as-a-Judge that evaluates the issued queries along two dimensions:
(1) \emph{query abstraction}, whether the query targets reusable principles or workflows rather than instance-specific details or an overly broad topic;
and (2) \emph{evidence gain}, whether the retrieved snippets resolve the question's key uncertainties with concrete, non-redundant support.
Poorly-posed queries incur a penalty $p_q(\tau) \in [0, 0.3]$ that discounts the quality bonus below,
so that a skill reached through bad queries earns little reward.

\paragraph{Skill-Generation Reward ($R_{\text{skill}}$).}
We add an execution-based \emph{reuse} score to a rubric-based LLM-as-a-Judge \emph{grounding} score:
(1) \emph{reuse} $s_{\text{reuse}} \in [0,1]$ measures the gain in the skill's execution accuracy on retrieved questions similar to $q$ relative to direct reasoning~\cite{ye2026umem};
and (2) \emph{grounding} $s_{\text{ground}} \in [0,1]$ judges whether the skill is faithfully grounded in the retrieved evidence rather than hallucinated from the model's own parametric memory.

\paragraph{Final Reward.}
The three decision rewards are composed through a gated structure:
the exploration-necessity term decides \emph{whether} a rollout earns a decision bonus,
and the search-quality and skill-generation terms decide \emph{how large} it is.
Formally, the unified decision reward of a trajectory $\tau$ is
\begin{equation}
s^{c}(\tau) =
\begin{cases}
q(\tau), & \Delta > \tau_0 \ \text{and}\ \tau \in G_{\text{exp}}, \\
1, & \Delta < -\tau_0 \ \text{and}\ \tau \in G_{\text{direct}}, \\
0, & \text{otherwise},
\end{cases}
\label{eq:decision}
\end{equation}
An exploring rollout earns the quality bonus $q(\tau)$ only when exploration clearly outperforms direct answering ($\Delta > \tau_0$);
symmetrically, a direct-answering rollout earns a unit bonus when exploration proves harmful ($\Delta < -\tau_0$);
otherwise no decision bonus is given.
This bonus $q(\tau)$ in Eq.~\eqref{eq:decision} averages the generated skill's reusability and grounding scores, discounted by the query penalty,
\begin{equation}
q(\tau) = \max\Big(0,\ \tfrac{1}{2}\big(s_{\text{reuse}}(\tau) + s_{\text{ground}}(\tau)\big) - p_q(\tau)\Big),
\label{eq:quality}
\end{equation}
so that the skill bonus is earned only through well-posed queries.
The final rollout reward then combines the task term and the decision reward with fixed weights,
\begin{equation}
R(\tau) = \operatorname{clip}_{[0,1]}\big(\lambda_a R_{\text{base}}(\tau) + \lambda_c s^{c}(\tau)\big),
\label{eq:reward}
\end{equation}
with $\lambda_a = 0.7$ and $\lambda_c = 0.3$.
By gating the quality reward behind a demonstrated capability gap, this composition naturally calibrates the agent's decision to search without encouraging indiscriminate exploration.

\begin{table*}[htbp]
\centering
\small
\setlength{\tabcolsep}{4pt}
\begin{tabular*}{\textwidth}{@{\extracolsep{\fill}}l ccccccc ccccccc@{}}
\toprule
\multirow{3}{*}{\textbf{Method}}
  & \multicolumn{7}{c}{\textbf{Qwen3-4B}}
  & \multicolumn{7}{c}{\textbf{Qwen3-8B}} \\
\cmidrule(lr){2-8}\cmidrule(lr){9-15}
  & \multicolumn{3}{c}{\textbf{SuperGPQA}} & \multicolumn{3}{c}{\textbf{MMLU-Pro}} & \multirow{2}{*}{\textbf{Avg}}
  & \multicolumn{3}{c}{\textbf{SuperGPQA}} & \multicolumn{3}{c}{\textbf{MMLU-Pro}} & \multirow{2}{*}{\textbf{Avg}} \\
\cmidrule(lr){2-4}\cmidrule(lr){5-7}\cmidrule(lr){9-11}\cmidrule(lr){12-14}
  & Math & Mgmt. & Sci. & Law & Phil. & Hist. &
  & Math & Mgmt. & Sci. & Law & Phil. & Hist. & \\
\midrule
\multicolumn{15}{l}{\textit{Search-Augmented Baselines}} \\
\midrule
Direct Inference  & 65.4 & 38.2 & 45.0 & 29.8 & 54.9 & 47.2 & 46.8\textsubscript{\phantom{+0.0}} & 64.8 & 41.4 & 45.6 & 39.5 & 58.3 & 58.8 & 51.4\textsubscript{\phantom{+0.0}}\\
Search Agent      & 64.0 & 37.8 & 47.8 & 28.5 & 56.7 & 50.1 & 47.5\textsubscript{+0.7} & 67.2 & 43.0 & 50.2 & 35.8 & 60.9 & 59.6 & 52.8\textsubscript{+1.4} \\
Search Agent\textsubscript{train} & 63.0 & \underline{42.2} & 58.8 & \textbf{39.5} & \underline{64.7} & \underline{55.9} & \underline{54.0}\textsubscript{+7.3} & 67.0 & 44.0 & 62.0 & \underline{43.6} & \underline{68.3} & 60.6 & \underline{57.7}\textsubscript{+6.3} \\
\midrule
\multicolumn{15}{l}{\textit{Skill-Learning Baselines}} \\
\midrule
ReasoningBank & 61.4 & 33.2 & 52.6 & 22.6 & 55.1 & 49.3 & 45.7\textsubscript{-1.1} & 64.6 & 38.0 & 48.8 & 25.0 & 55.9 & 57.7 & 48.3\textsubscript{-3.1} \\
Memp          & 62.2 & 37.4 & 51.8 & 21.2 & 54.7 & 51.4 & 46.5\textsubscript{-0.3} & 67.2 & 41.0 & 45.6 & 27.7 & 60.9 & 57.5 & 49.9\textsubscript{-1.5} \\
EvolveR\textsubscript{train} & 61.2 & 39.8 & 
\underline{59.2} & 33.3 & 59.1 & 55.4 & 51.3\textsubscript{+4.5} & 64.0 & 46.0 & \textbf{63.8} & 41.2 & 67.5 & \underline{61.9} & 57.4\textsubscript{+6.0} \\

\midrule
\multicolumn{15}{l}{\textit{Search2Skill (Ours)}} \\
\midrule
Search2Skill          & \underline{65.6} & 39.0 & 50.4 & 28.8 & 56.9 & 50.4 & 48.5\textsubscript{+1.7} & \underline{68.0} & \underline{46.0} & 48.0 & 38.4 & 62.3 & 59.6 & 53.7\textsubscript{+2.3} \\
\textbf{Search2Skill\textsubscript{train}} & \textbf{67.2} & \textbf{43.0} & \textbf{59.4} & \underline{36.6} & \textbf{67.1} & \textbf{57.0} & \textbf{55.0}\textsubscript{+8.3} & \textbf{71.6} & \textbf{50.0} & \underline{63.2} & \textbf{45.4} & \textbf{68.7} & \textbf{65.1} & \textbf{60.7}\textsubscript{+9.3} \\
\bottomrule
\end{tabular*}
\caption{\textbf{Streaming setting.} We report average accuracy over three random question orders, for Qwen3-4B and Qwen3-8B backbones.
\emph{Mgmt.}, \emph{Sci.}, \emph{Phil.}, and \emph{Hist.}\ denote Management, Science, Philosophy, and History, respectively.
}
\label{tab:online}
\end{table*}

\begin{table*}[t]
\centering
\small
\setlength{\tabcolsep}{2.5pt}
\begin{tabular*}{\textwidth}{@{\extracolsep{\fill}}l cccccccc cccccccc@{}}
\toprule
\multirow{3}{*}{\textbf{Method}}
  & \multicolumn{8}{c}{\textbf{Qwen3-4B}}
  & \multicolumn{8}{c}{\textbf{Qwen3-8B}} \\
\cmidrule(lr){2-9}\cmidrule(lr){10-17}
  & \multicolumn{2}{c}{\textbf{EvoAgentBench}} & \multicolumn{3}{c}{\textbf{MMLU-Pro}} & \multicolumn{2}{c}{\textbf{SuperGPQA}} & \multirow{2}{*}{\textbf{Avg}}
  & \multicolumn{2}{c}{\textbf{EvoAgentBench}} & \multicolumn{3}{c}{\textbf{MMLU-Pro}} & \multicolumn{2}{c}{\textbf{SuperGPQA}} & \multirow{2}{*}{\textbf{Avg}} \\
\cmidrule(lr){2-3}\cmidrule(lr){4-6}\cmidrule(lr){7-8}\cmidrule(lr){10-11}\cmidrule(lr){12-14}\cmidrule(lr){15-16}
  & Math & Code & Law & Phil. & Hist. & Math & Sci. &
  & Math & Code & Law & Phil. & Hist. & Math & Sci. & \\
\midrule
Direct Inference            & 41.0 & 62.5 & 27.4 & 53.6 & 52.8 & 58.7 & 52.5 & 49.8 & 45.0 & 64.1 & 33.1 & 51.8 & 57.5 & 67.3 & 50.0 & 52.7 \\
\midrule
\multicolumn{17}{l}{\textit{Skill-Learning Baselines}} \\
\midrule
ReasoningBank               & 36.0 & 61.5 & 31.5 & 51.8 & 51.2 & 59.3 & 52.0 & 49.0\textsubscript{-0.8} & 43.0 & 64.1 & 33.9 & 52.4 & 57.5 & 65.3 & 54.0 & 52.9\textsubscript{+0.2} \\
Memp                        & 38.0 & 64.1 & \textbf{33.9} & 51.8 & 54.1 & 62.7 & \textbf{58.0} & 51.8\textsubscript{+2.0} & 44.0 & 66.7 & 36.3 & 56.6 & \underline{59.1} & 66.0 & \textbf{56.5} & 55.0\textsubscript{+2.3} \\
Trace2Skill                 & 35.0 & \underline{67.9} & 29.0 & \underline{56.0} & \underline{54.3} & 60.7 & 53.5 & 50.9\textsubscript{+1.1} & \underline{49.0} & \underline{68.0} & 37.9 & \underline{57.2} & 57.5 & 63.3 & 53.0 & 55.1\textsubscript{+2.4} \\
SkillOpt                    & 37.0 & 61.5 & 27.4 & 50.1 & 52.8 & 60.7 & 49.5 & 48.4\textsubscript{-1.4} & 43.0 & 67.0 & 32.1 & 49.8 & 55.1 & 69.3 & 51.7 & 52.6\textsubscript{-0.1} \\
SkillOpt (32B)              & 39.0 & 64.1 & 28.2 & \underline{56.0} & 52.3 & \underline{64.0} & 52.5 & 50.9\textsubscript{+1.1} & \underline{49.0} & 67.0 & 33.1 & 54.2 & \underline{59.1} & \underline{71.3} & 52.3 & 55.1\textsubscript{+2.4} \\
\midrule
\multicolumn{17}{l}{\textit{Search2Skill (Ours)}} \\
\midrule
Search2Skill                & \underline{41.0} & 64.1 & 25.8 & 50.6 & 51.6 & 62.3 & 51.0 & 49.5\textsubscript{-0.3} & 45.0 & 64.1 & \underline{39.9} & 56.0 & 55.1 & 69.3 & 54.0 & 54.8\textsubscript{+2.1} \\
\textbf{Search2Skill\textsubscript{train}} & \textbf{43.0} & \textbf{68.4} & \underline{32.4} & \textbf{57.2} & \textbf{59.8} & \textbf{67.3} & \underline{56.5} & \textbf{54.9}\textsubscript{+5.1} & \textbf{52.0} & \textbf{71.8} & \textbf{41.9} & \textbf{59.0} & \textbf{62.2} & \textbf{72.7} & \underline{55.5} & \textbf{59.3}\textsubscript{+6.6} \\
\bottomrule
\end{tabular*}
\caption{\textbf{Held-out setting.} Skills are mined on a collection split and reused on held-out test questions with search disabled; all methods  share the same executor model (Qwen3-4B or Qwen3-8B).
}
\label{tab:reuse}
\end{table*}

\section{Experimental Setup}\label{sec:setup}

We describe the training procedure, evaluation benchmarks, two complementary protocols, and compared baselines,
followed by the implementation details.

\subsection{Training Details}
The policy is first cold-started by supervised fine-tuning on 8K high-quality agent trajectories generated by DeepSeek-V3.2~\cite{liu2025deepseek} and GLM-5~\cite{zeng2026glm} models on open-source questions~\cite{ma2026general},
and then optimized with GRPO~\cite{shao2024deepseekmath} on 2K filtered high-quality questions spanning five domains (Physics, Chemistry, Finance, Business, and Economics).

\subsection{Benchmarks}
We evaluate Search2Skill on eight domains from three expert-level benchmarks:
SuperGPQA~\cite{du2026supergpqa} (Math, Management, Science),
MMLU-Pro~\cite{wang2024mmlu} (Law, Philosophy, History),
and EvoAgentBench~\cite{gao2026evoagentbench} (OmniMath and LiveCodeBench).\footnote{EvoAgentBench contains five splits in total; the remaining three require sandboxed execution, while our training data does not involve sandbox-based agentic RL.}
Most of these domains are not covered by the training data and thus test out-of-domain generalization.
Each domain provides up to $500$ questions: SuperGPQA domains are filtered to medium and hard difficulty, while MMLU-Pro domains with fewer than $500$ qualified questions are used in full.

\subsection{Evaluation Protocol}
We design two complementary protocols:
(1) \textbf{Streaming protocol} processes questions as a sequential stream~\cite{ye2026umem}:
skills generated from earlier questions are immediately written into the skill library and become available to subsequent ones,
measuring whether the agent continuously accumulates reusable knowledge. We report streaming accuracy averaged over three random question orders;
(2) \textbf{Held-out protocol} partitions the questions of each domain into a \emph{collection} set for skill mining and a held-out \emph{test} set (2:1 ratio),
and reuses the frozen skill library on the test questions, measuring intrinsic skill reusability~\cite{gao2026evoagentbench}.

\subsection{Baselines}
We compare with two categories of baselines.
(1) \textbf{Skill-learning baselines}:
ReasoningBank~\cite{ouyang2025reasoningbank} and Memp~\cite{fang2026memp} distill experience from past trajectories via memory-based retrieval;
Trace2Skill~\cite{ni2026trace2skill} refines a global skill document offline;
SkillOpt~\cite{yang2026skillopt} iteratively optimizes a skill document from rollout feedback, using the executor itself or Qwen3-32B as the optimizer;
and EvolveR\textsubscript{train}~\cite{wu2026evolverselfevolvingllmagents} is an RL-trained self-evolving memory method, re-trained under our setup for fair comparison.
EvolveR is evaluated only under the streaming protocol, as its experience guides search behavior and is incompatible with the held-out setting without searching, whereas Trace2Skill and SkillOpt optimize at the batch level and are evaluated only under the held-out protocol;
and (2) \textbf{Search-augmented baselines}, included to isolate raw search capability from persistent skill learning:
Direct Inference answers without search;
Search Agent equips the base model with the same search tools without training;
and Search Agent\textsubscript{train} is a search agent trained with GRPO on the same training data, following deep search paradigms~\cite{jin2025search,liu2025webexplorer,li2025deepagent,zhu2026marco} but without any skill-learning mechanism.

\subsection{Implementation Details}
All search-based methods use the same agent environment: a Google Search API tool returning the Top-5 URLs per query, a page-visit tool that compresses fetched pages with Qwen3-30B-A3B, and a sandboxed Python tool~\cite{zhu2026marco}. More details on datasets, configurations, and hyperparameters are provided in the Appendix.

\section{Main Experimental Results}\label{sec:main_exp}

We evaluate Search2Skill under the two complementary protocols above:
the \emph{streaming} protocol measures continuous skill accumulation over a task stream,
and the \emph{held-out} protocol measures the intrinsic reusability of the acquired skills.

\subsection{Streaming Evaluation: Dynamic Skill Distillation}

Table~\ref{tab:online} reveals that $\text{Search2Skill}_{\text{train}}$ achieves the best average accuracy on both backbones, improving over Direct Inference by +8.3\% on Qwen3-4B and +9.3\% on Qwen3-8B.
Since this setting mirrors realistic self-evolution, where the agent accumulates skills from a continuous question stream while reasoning over incoming tasks,
these gains demonstrate stronger continual learning ability than all baselines.
We analyze these gains from three perspectives below.

\paragraph{Comparison with skill-learning baselines.}
Table~\ref{tab:online} reveals that ReasoningBank and Memp even fall below Direct Inference on average on both backbones.
By contrast, Search2Skill fills the knowledge gap through external search and improves consistently on both backbones. Furthermore, Search2Skill outperforms EvolveR\textsubscript{train}, which is also RL-trained under our setup, by +3.7\% on Qwen3-4B and +3.3\% on Qwen3-8B, highlighting the importance of distilling skills from searched evidence.

\paragraph{Comparison with search-augmented baselines.}
The gain does not come merely from adding search tools:
$\text{Search2Skill}_{\text{train}}$ outperforms the trained search baseline $\text{Search Agent}_{\text{train}}$ by +1.0\% on 4B and +3.0\% on 8B,
showing that Search2Skill better calibrates when to search and how to distill skills, instead of searching more evidence.

\paragraph{Per-domain gains.}
The improvements over Direct Inference hold in every domain on both backbones.
Even on the few domains where $\text{Search Agent}_{\text{train}}$ and $\text{EvolveR}_{\text{train}}$ are slightly stronger (Law on Qwen3-4B and Science on Qwen3-8B), the gaps are small (2.9\% and 0.6\% points, respectively),
and Search2Skill achieves the best average accuracy overall.

\subsection{Held-Out Evaluation: Intrinsic Skill Reusability}
Table~\ref{tab:reuse} shows that, without web access on held-out test splits,
$\text{Search2Skill}_{\text{train}}$ improves over Direct Inference by +5.1\% on Qwen3-4B and +6.6\% on Qwen3-8B, while outperforming most trajectory-based baselines.

\paragraph{Skill abstraction versus raw evidence.}
A natural question is whether the above gains come merely from web search, i.e., whether caching the retrieved content alone yields comparable gains.
Table~\ref{tab:web_vs_skill} rules this out: reusing raw retrieved evidence improves over Direct Inference by only +1.8\% on average,
whereas reusing the abstracted skills distilled from the same evidence yields +6.3\%, a +4.5\% margin of abstraction over raw caching.
This confirms that Search2Skill compresses scattered web evidence into reusable procedural knowledge, rather than simply storing retrieved content.

\begin{table}[t]
\centering
\small
\begin{tabular*}{\columnwidth}{@{\extracolsep{\fill}}lcccc@{}}
\toprule
\textbf{Stored Artifact} & \textbf{Law} & \textbf{History} & \textbf{Math} & \textbf{Avg} \\
\midrule
Direct Inference & 33.1 & 57.5 & 67.3 & 52.6\textsubscript{\phantom{+0.0}} \\
Raw retrieved evidence & 36.3 & 58.3 & 68.7 & 54.4\textsubscript{+1.8} \\
Abstracted skill       & \textbf{41.9} & \textbf{62.2} & \textbf{72.7} & \textbf{58.9}\textsubscript{+6.3} \\
\bottomrule
\end{tabular*}
\caption{\textbf{Raw evidence vs.\ skill (Qwen3-8B).} Both forms reuse the same information gathered during past search.}
\label{tab:web_vs_skill}
\end{table}

\begin{figure}[t]
\centering
\includegraphics[width=\linewidth]{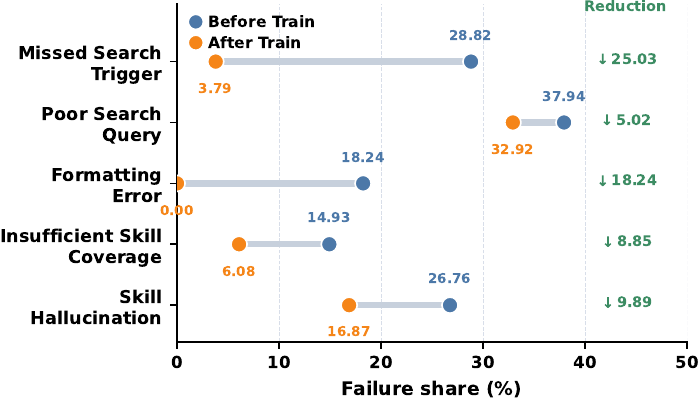}
\caption{\textbf{Failure patterns before and after RL training.} RL reduces the failure share of all five patterns.}
\label{fig:error_analysis}
\end{figure}

\paragraph{Substantial improvement after training.}
Without RL training, the naive Search2Skill transfers poorly and even falls below Direct Inference on Qwen3-4B.
Our error analysis in Analysis Section attributes this to two failure modes:
missed search triggering on questions with knowledge gaps, and skill hallucination induced by noisy web evidence (Figure~\ref{fig:error_analysis}).
Rubric-RL mitigates both, turning the negative delta into net gains of +5.1\% on 4B and +6.6\% on 8B over Direct Inference.

\paragraph{Cross-scale capacity support.}
Search2Skill skills could transfer across model scales:
Qwen3-4B with Search2Skill reaches 56.5\% on Science, surpassing the unassisted Qwen3-8B Direct Inference at 50.0\%.
Structured skills thus act as an external capability scaffold, allowing a smaller model to compete with a larger one.
The net gain over Direct Inference is also larger on 8B than on 4B, suggesting that stronger backbones can exploit skill guidance more effectively.

\section{Analysis}
\label{sec:analysis}

The main experiments show that Search2Skill improves downstream performance;
this section investigates \emph{why} it works through four research questions:
whether RL resolves the dominant failure patterns of the untrained model (RQ1),
whether Search2Skill acquires skills more efficiently than existing methods (RQ2),
whether each rubric-based reward contributes (RQ3),
and whether the learned skills transfer across executor models (RQ4).

\subsection{RQ1: Does RL Resolve the Failure Patterns?}

Beyond final accuracy, we examine how RL reshapes the agent's behavior.
We define five failure patterns along the Search2Skill pipeline (Figure~\ref{fig:error_analysis}):
(1) \textbf{Missed search trigger}, where the agent fails to search when needed;
(2) \textbf{Poor search query}, where issued queries are misdirected or overly generic;
(3) \textbf{Formatting error}, i.e., invalid actions such as looping over search without producing a skill or an answer;
(4) \textbf{Insufficient skill coverage}, where no suitable skill exists or is retrieved;
and (5) \textbf{Skill hallucination}, where generated skills fabricate or distort the retrieved evidence.
We prompt a teacher model to identify these patterns from the trajectories of $500$ evaluation questions,
and measure each pattern's failure share, i.e., the fraction of questions exhibiting it:
missed search trigger is measured over all questions, since skipping search can occur on any question,
while the remaining patterns are measured over questions where search is triggered.
Figure~\ref{fig:error_analysis} shows that RL reduces the failure share of every pattern, from 25.3\% to 11.9\% on average,
indicating that RL improves the full pipeline rather than fixing a single stage.

\subsection{RQ2: Is Search2Skill More Efficient?}

\begin{figure}[t]
    \centering
    \includegraphics[width=\linewidth]{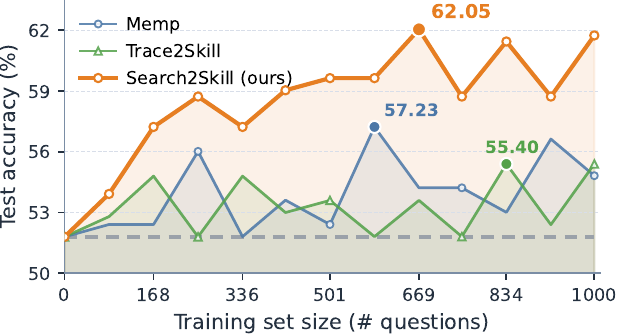}
    \caption{\textbf{Skill acquisition efficiency on MMLU-Pro Philosophy.} Search2Skill reaches higher accuracy with fewer training questions than baselines.}
    \label{fig:skill_scaling}
\end{figure}

We evaluate skill acquisition efficiency on MMLU-Pro Philosophy by having each method perform three evolution passes over the collection set, with intermediate checkpoints evaluated on the test set (Figure~\ref{fig:skill_scaling}).
Overall, the inward-looking baselines fluctuate around their initial accuracy and barely improve over the full run:
confined by the model's own capability boundary, they acquire useful skills at a low rate.
Search2Skill instead breaks this stagnation and shows three advantages:
\textbf{(1) Rapid early acquisition:} accuracy jumps from 51.8\% to 57.2\% after only 168 questions, while Memp and Trace2Skill remain largely stagnant, confirming that searching outward resolves capability gaps immediately;
\textbf{(2) Superior sample efficiency:} Search2Skill reaches 59.6\% halfway through training, whereas no baseline reaches 57\% even after full training;
and \textbf{(3) Consistent accumulation:} Search2Skill improves steadily, while Trace2Skill oscillates severely due to noisy inward-summarized skills.

\subsection{RQ3: Do Rubric-Based Rewards Contribute?}

\begin{table}[t]
\centering
\small
\begin{tabular*}{\columnwidth}{@{\extracolsep{\fill}}lccccc@{}}
\toprule
\textbf{Method} & \textbf{Math} & \textbf{History} & \textbf{Law} & \textbf{Avg} & \begin{tabular}{@{}c@{}}\textbf{Search}\\\textbf{Rate}\end{tabular} \\
\midrule
Search2Skill & \textbf{71.6} & \textbf{65.1} & \textbf{45.4} & \textbf{60.7}\textsubscript{\phantom{-0.0}} & 65.9\% \\
\midrule
\quad w/o $r_{\mathrm{when}}$ & 66.2 & 61.9 & 42.5 & 56.9\textsubscript{-3.8} & 96.6\% \\
\quad w/o $r_{\mathrm{sea}}$ & 70.0 & 63.2 & 43.5 & 58.9\textsubscript{-1.8} & 70.2\% \\
\quad w/o $r_{\mathrm{gen}}$ & 68.0 & 60.4 & 44.5 & 57.6\textsubscript{-3.1} & 72.1\% \\
\quad w/o all & 67.1 & 62.4 & 40.7 & 56.7\textsubscript{-4.0} & 57.8\% \\
\bottomrule
\end{tabular*}
\caption{\textbf{Reward ablation.} $r_{\mathrm{when}}$, $r_{\mathrm{sea}}$, and $r_{\mathrm{gen}}$ denote the \emph{exploration-necessity}, \emph{search-quality}, and \emph{skill-generation} rewards; the last row keeps only the task reward $R_{\text{base}}$.}
\label{tab:reward_ablation}
\end{table}

We remove each reward in turn and test a variant trained with only the correctness reward (Table~\ref{tab:reward_ablation}).
All variants degrade, most severely when only the base reward is kept ($-4.0\%$),
The exploration-necessity reward $r_{\mathrm{when}}$ contributes the most:
without it, the search rate surges to $96.6\%$ and average accuracy drops by $3.8\%$.
Trajectory inspection shows that uncontrolled search wastes tokens on easy questions and produces simple, low-quality skills, which further hurt subsequent questions.
These results confirm that the three rubric rewards provide complementary supervision.

\subsection{RQ4: Do the Skills Transfer Across Executors?}

The controlled comparison in the main experiments shows that abstracted skills outperform the raw evidence they were distilled from (Table~\ref{tab:web_vs_skill}).
A further question is whether these skills are tied to the model that collected them.

We reuse the skill library mined by the RL-trained 8B model with Qwen3-4B and Qwen3-14B executors, without any further skill generation (Table~\ref{tab:cross_model}).
The 8B-mined library improves over Direct Inference by +4.1\% on the 4B executor and +3.5\% on the 14B executor, outperforming all corresponding baselines.
Notably, these skills are distilled from questions that the 8B model cannot solve by itself,
yet the missing knowledge they encode is not specific to that model.
This confirms that Search2Skill captures reusable procedural knowledge rather than model-specific behaviors.

\begin{table}[htbp]
\centering
\small
\setlength{\tabcolsep}{6pt}
\begin{tabular*}{\columnwidth}{@{\extracolsep{\fill}}lcccc@{}}
\toprule
\textbf{Executor / Method} & \textbf{Law} & \textbf{History} & \textbf{Math} & \textbf{Avg.} \\
\midrule
\multicolumn{5}{l}{\textit{Qwen3-4B executor}} \\
\midrule
Direct                & 27.4 & 52.8 & 58.7 & 46.3\textsubscript{\phantom{+0.0}} \\
ReasoningBank         & 31.5 & 51.2 & 59.3 & 47.3\textsubscript{+1.0} \\
Memp                  & \textbf{33.9} & 54.1 & 62.7 & 50.2\textsubscript{+3.9} \\
Trace2Skill           & 29.0 & 54.3 & 60.7 & 48.0\textsubscript{+1.7} \\
Search2Skill Skills   & 32.4 & \textbf{54.8} & \textbf{64.1} & \textbf{50.4}\textsubscript{+4.1} \\
\midrule
\multicolumn{5}{l}{\textit{Qwen3-14B executor}} \\
\midrule
Direct                & 41.1 & 59.1 & 76.0 & 58.7\textsubscript{\phantom{+0.0}} \\
ReasoningBank         & 41.1 & 60.6 & 68.7 & 56.8\textsubscript{-1.9} \\
Memp                  & \textbf{45.2} & 58.3 & 68.7 & 57.4\textsubscript{-1.3} \\
Trace2Skill           & 43.5 & 58.3 & 75.5 & 59.1\textsubscript{+0.4} \\
Search2Skill Skills   & \textbf{45.2} & \textbf{62.2} & \textbf{79.3} & \textbf{62.2}\textsubscript{+3.5} \\
\bottomrule
\end{tabular*}
\caption{\textbf{Cross-model transfer of Search2Skill skills.}
A skill library mined once by the RL-trained 8B collector is reused by different executors without further skill generation. }
\label{tab:cross_model}
\end{table}

\section{Conclusion}\label{sec:conclusion}
In this work, we presented Search2Skill, a search-driven framework that turns capability gaps identified during reasoning into reusable skills, optimized with a rubric-based reinforcement learning scheme that governs when to search, how to search, and how to distill skills.
Experiments on eight expert-level domains show that Search2Skill outperforms both skill-learning and search-augmented baselines under streaming and held-out evaluation protocols.
Further analyses show that RL repairs failure patterns across the full pipeline with complementary supervision from each rubric reward, and that the gains stem from skill abstraction rather than raw retrieved evidence. These results suggest that external exploration paves the way for self-evolving agents that grow beyond capability boundaries.

\bibliography{references}

\clearpage
\appendix
\setcounter{secnumdepth}{2}

\noindent This technical appendix covers the evaluation setup, the training and reward details, the implementation of the preliminary study, the failure-pattern analysis, run-to-run variance, the prompt templates that drive Search2Skill's agent loop, and example trajectories and acquired skills.

\section{Evaluation Setup}\label{app:eval}

We describe the evaluation benchmarks, dataset construction, agent environment, and skill retrieval procedure used in the experiments reported in the main paper.

\subsection{Benchmarks and Dataset Statistics}\label{app:benchmarks}

\paragraph{Benchmarks.}
We evaluate on three benchmarks. SuperGPQA~\cite{du2026supergpqa} and MMLU-Pro~\cite{wang2024mmlu} are large-scale, expert-level multiple-choice benchmarks covering diverse professional disciplines. From SuperGPQA, we use the \emph{Math}, \emph{Management}, and \emph{Science} domains; from MMLU-Pro, we use \emph{Law}, \emph{Philosophy}, and \emph{History}, some of which contain fewer than 500 qualified questions and are therefore used in full. We also evaluate on EvoAgentBench~\cite{gao2026evoagentbench}, a benchmark designed to test skill reuse, from which we select its two QA-style datasets, \emph{OmniMath} and \emph{LiveCodeBench}, used only in the reuse-focused Held-out setting described below. The remaining three splits require rollout-time interaction with a stateful file-system sandbox, which lies outside our agent's action space. Because the LiveCodeBench test split contains only 39 tasks, we evaluate it three times and report the mean.

\paragraph{Dataset sizes.}
Table~\ref{tab:app-data} summarizes the dataset sizes under both evaluation protocols. For SuperGPQA, we randomly sample up to 500 questions from the selected domains after filtering to the medium and hard difficulty levels. For MMLU-Pro, some domains contain fewer than 500 qualified questions, so we use all available instances. In the \textbf{Held-out} setting, SuperGPQA and MMLU-Pro are randomly split into collection and test sets at a strict 2:1 ratio, while EvoAgentBench follows its official benchmark splits.

\paragraph{Correctness evaluation.}
We report accuracy on all benchmarks, but the correctness of an answer is judged differently across benchmarks. For the multiple-choice benchmarks SuperGPQA and MMLU-Pro, we extract the option letter from the agent's \texttt{<ANSWER>} tag and require an exact match with the gold option. For EvoAgentBench, OmniMath answers are open-ended, so we use Qwen3-30B-A3B as a judge model to decide whether the agent's final answer is equivalent to the reference answer; LiveCodeBench solutions are executed against the unit tests shipped with the dataset, and a solution counts as correct only if it passes all of them.

\begin{table}[t]
\centering
\footnotesize
\setlength{\tabcolsep}{4.5pt}
\begin{tabular}{@{}ll c cc@{}}
\toprule
\multirow{2}{*}{\textbf{Benchmark}} & \multirow{2}{*}{\textbf{Domain}}
  & \textbf{Streaming} & \multicolumn{2}{c}{\textbf{Held-out}} \\
\cmidrule(lr){3-3}\cmidrule(lr){4-5}
  & & \#Q & Collect. & Test \\
\midrule
\multirow{3}{*}{SuperGPQA}
  & Mathematics & 500 & 333 & 167 \\
  & Management  & 500 & --  & --  \\
  & Science     & 500 & 333 & 167 \\
\midrule
\multirow{3}{*}{MMLU-Pro}
  & Law        & 372 & 248 & 124 \\
  & Philosophy & 500 & 333 & 167 \\
  & History    & 381 & 254 & 127 \\
\midrule
\multirow{2}{*}{EvoAgentBench}
  & OmniMath      & -- & 478 & 100 \\
  & LiveCodeBench & -- &  97 &  39 \\
\bottomrule
\end{tabular}
\caption{Per-domain dataset sizes for the Streaming and Held-out evaluation protocols.}

\label{tab:app-data}
\end{table}

\subsection{Agent Environment}\label{app:env}
During both data collection and online evaluation, the agent's exploration loop relies on three tools: (1)~a \textbf{web search} tool backed by the Google Search API, which returns the top-5 most relevant URLs for each query; (2)~a \textbf{page visit} tool that fetches a selected URL and compresses its content via Qwen3-30B-A3B to extract task-relevant information before returning it to the agent; and (3)~a \textbf{code} tool that runs the agent's code in a sandboxed Python interpreter, exposing the executable functions defined in the \texttt{code} field of activated skills so that the agent can invoke a reusable skill function and obtain its result rather than performing the computation by hand. To prevent the agent from simply retrieving the exact problem and its solution, we further add an \emph{original-question blocking} mechanism: during page visit we detect whether the fetched page contains the original question stem and discard it if so, and we additionally exclude common benchmark-hosting sites such as Hugging Face and GitHub. This forces the agent to learn the underlying knowledge and method rather than copying the original question and answer.

\paragraph{Inference configuration.}
All methods decode with temperature $0.3$ and Qwen3's thinking mode enabled. Each episode is limited to $15$ interaction rounds and at most $5$ search calls; an episode that exhausts this budget without emitting an \texttt{<ANSWER>} action is counted as a failure.

\subsection{Skill Retrieval}\label{app:retrieval}

Each skill follows the standardized schema of Prompt~2: \texttt{skill\_name}, \texttt{use\_when} (an applicability description), \texttt{workflow}, and \texttt{code}. In addition, we store the source question from which it was extracted as retrieval-only metadata (it is not part of the generated skill). We represent each skill by its textual description $d_i$ (the \texttt{skill\_name} and \texttt{use\_when} fields) together with its source question $\tilde{q}_i$.

Given a task question $q$, we encode the query, skill descriptions, and source questions using the fixed embedding model \texttt{text-embedding-v4} (1024 dimensions). The relevance of a skill is computed as
\begin{equation}
\begin{split}
\mathrm{score}(q,s_i) = {}
&\gamma\,\cos\!\big(\phi(q),\phi(d_i)\big) \\
&+ (1-\gamma)\,\cos\!\big(\phi(q),\phi(\tilde{q}_i)\big),
\end{split}
\label{eq:app-retrieval}
\end{equation}
where $\phi(\cdot)$ denotes the embedding model, $d_i$ is the skill description, and $\tilde{q}_i$ is the corresponding source question. We use $\gamma=0.7$, retrieve the top-$k=3$ skills, and discard skills whose similarity is below $\theta=0.50$.

Retrieved skills are initially presented to the agent only through their names and applicability descriptions. The detailed workflow and executable code are revealed only after the agent explicitly activates the selected skill.

\paragraph{Library maintenance.}
The library is capped at $300$ skills, with no merging or rewriting of entries. The only update path is name collision: when a retrieved skill fails to solve the current question, the agent may distill a new one, and if it carries an existing \texttt{skill\_name} it overwrites the old entry. Any improvement in library quality therefore comes from better skill generation rather than from post-hoc library curation.

\subsection{Baseline Implementation Details}\label{app:baselines}

All baselines use the same executor backbone, agent environment (Appendix~\ref{app:env}), and embedding model (\texttt{text-embedding-v4}) as Search2Skill.

\paragraph{Direct Inference.}
The backbone answers each question with chain-of-thought reasoning, without tools or memory.

\paragraph{Search Agent \& Search Agent\textsubscript{train}.}
The backbone equipped with the same search, page-visit, and Python tools in a ReAct-style loop~\cite{jin2025search}, where each question is solved independently and nothing persists across questions. The \emph{train} variant is further trained with GRPO on our 2K RL questions using only the task reward $R_{\text{base}}$, following deep search paradigms~\cite{liu2025webexplorer,li2025deepagent,zhu2026marco}, isolating RL-trained search capability from skill learning. Since they keep no reusable artifact and the held-out protocol forbids searching again, both variants are evaluated only under the streaming protocol.

\paragraph{ReasoningBank~\cite{ouyang2025reasoningbank}.}
A memory-augmented method that distills reusable reasoning insights from past trajectories (both successes and failures) into a searchable bank; the top-$3$ relevant items are retrieved and injected into the executor's context. It represents experience-distillation approaches without external search.

\paragraph{Memp~\cite{fang2026memp}.}
A procedural-memory method that induces reusable procedures from agent experience and maintains the store via rule-based consolidation and updating, retrieved by question-embedding similarity (top-$3$).

\paragraph{Trace2Skill~\cite{ni2026trace2skill}.}
An offline framework for skill evolution that splits trajectories into success and failure cases, proposes trajectory-specific skill patches with parallel analyst agents, and holistically merges them into a single comprehensive skill directory. The resulting skill is used directly at inference by prepending its root document to the prompt. Evaluated only under the held-out protocol due to its batch-level optimization.

\paragraph{SkillOpt~\cite{yang2026skillopt}.}
A text-space skill optimization method for frozen agents that treats the skill document as a trainable external text state: an independent optimizer model proposes controlled add/delete/edit operations from execution trajectories, and a validation-set gate retains only updates that improve performance. Since we find that using the 4B/8B backbones themselves as the optimizer is weak at such editing, we additionally report \emph{SkillOpt (32B)}, which replaces the optimizer with Qwen3-32B while keeping rollout collection and execution on the original 4B/8B backbone. Evaluated only under the held-out protocol.

\paragraph{EvolveR\textsubscript{train}~\cite{wu2026evolverselfevolvingllmagents}.}
A closed-loop self-evolving framework that distills past trajectories into reusable strategic principles, stores them in a dynamically maintained experience base, and retrieves them to guide future reasoning and search. Although EvolveR also couples search with evolving memory, its memory is induced from the agent's own trajectories and mainly improves reuse of internal experience, whereas Search2Skill searches external sources to fill identified capability gaps and distills the retrieved evidence into skills, enabling growth beyond what trajectories alone can provide. We re-train both the 4B and 8B backbones with their method on our own RL data under the same hyperparameters as Appendix~\ref{app:rl}. Since its experience steers live search, it is evaluated only under the streaming protocol.

\section{Training Details}\label{app:train}

\paragraph{Training Data Source.}
We derive all training data from WebInstruct-verified~\cite{ma2026general}, a large-scale corpus of 229K instructional question--answer pairs crawled from educational and professional websites and subsequently verified for correctness. WebInstruct covers a broad range of domains including science, engineering, business, and medicine, making it well-suited for training agents that must acquire expert-domain skills. We use this corpus as the problem source for both the SFT and RL stages. After our filtering pipeline, the retained questions concentrate mainly in five domains: Physics, Chemistry, Finance, Business, and Economics.

\subsection{Supervised Fine-Tuning}\label{app:sft}

\paragraph{Data Collection.}
We collect SFT demonstrations by running two teacher models---DeepSeek-V3.2~\cite{liu2025deepseek} and GLM-5~\cite{zeng2026glm}---on questions sampled from WebInstruct-verified. Each teacher executes the full Search2Skill agent loop autonomously, deciding whether to explore or answer directly. We apply rejection sampling to retain only the trajectories on which the teacher reaches the correct final answer, so that the policy is imitated from verified-correct agent behavior. This yields approximately 8K high-quality trajectories covering both exploration and direct-answer behaviors, used for both Qwen3-4B and Qwen3-8B.

\paragraph{Training Configuration.}
We perform full-parameter fine-tuning on Qwen3-8B with the following hyperparameters: learning rate $1\times10^{-5}$ with cosine decay, 10 warmup steps, global batch size 64, maximum sequence length 20{,}480 tokens, weight decay 0.1, AdamW ($\beta_1{=}0.9, \beta_2{=}0.95$), bf16 precision, and DeepSpeed ZeRO Stage-3. The Qwen3-4B policy is cold-started with exactly the same data and hyperparameters. Training runs for 3 epochs on 16 NVIDIA A100 GPUs and completes in approximately 1.5 hours for Qwen3-8B and 1 hour for Qwen3-4B.

\subsection{Reinforcement Learning}\label{app:rl}

\paragraph{Data Construction.}
We construct the RL training set through a filtering pipeline designed to select questions that genuinely require exploration.
We first compute sentence-embedding similarities among WebInstruct-verified questions and select anchor questions that possess sufficient semantic neighbors, ensuring that each anchor is accompanied by related questions for skill-reuse evaluation.
For each anchor, a teacher model Qwen3.7-Max~\cite{qwen37blog} produces a verified ground-truth answer. We then test the student model (Qwen3-8B) on each anchor; only questions that the student consistently fails are retained. Finally, the teacher relabels the ground truth of each anchor's similar neighbors.
The resulting dataset contains approximately 2K anchor questions, each paired with 3 similar questions used for online skill-reuse evaluation during RL training.
Qwen3-4B uses the same RL training data as Qwen3-8B.

\paragraph{Training Configuration.}
We optimize the SFT-initialized policy using GRPO~\cite{shao2024deepseekmath} implemented in the verl framework. Key hyperparameters: learning rate $1\times10^{-6}$, KL regularization coefficient $\beta{=}0.001$ (low-variance KL), train batch size 64, rollout group size $N{=}16$, maximum prompt length 2{,}048 tokens for the initial question prompt, maximum response length 18{,}431 tokens covering the whole multi-round rollout including tool observations, gradient checkpointing enabled, FSDP for the actor and reference model offloading. Rollouts are generated asynchronously via sglang with tensor parallelism. Tool observations (search results, visited page content, and format feedback) remain in the context but are masked out of the loss, so the policy-gradient and KL terms are computed only on the tokens generated by the policy itself. The reward is summarized in the main paper and specified in full in Appendix~\ref{app:reward}. The Qwen3-4B and Qwen3-8B policies are trained with identical hyperparameters; training runs for 3 epochs on 8 NVIDIA A100 GPUs and completes in approximately 17 and 22 hours, respectively.

\section{Reward Design Details}\label{app:reward}

The main paper decomposes the Search2Skill reward into a task term $R_{\text{base}}$ and three light-weight, rubric-based rewards for the core decisions of the framework: the exploration-necessity reward $R_{\text{when}}$ (\emph{when to search}), the search-quality reward $R_{\text{search}}$ (\emph{how to search}), and the skill-generation reward $R_{\text{skill}}$ (\emph{how to generate skills}). We keep the design deliberately simple: the group-contrastive gate of $R_{\text{when}}$ decides \emph{whether} a rollout earns a decision bonus, and a single quality magnitude $q_i$, composed from $R_{\text{search}}$ and $R_{\text{skill}}$, decides \emph{how large} it is. For a rollout group $\mathcal{G}$ of size $N{=}16$, rollout $i$ receives
\begin{equation}
R_i \;=\; \operatorname{clip}_{[0,1]}\!\big(\lambda_a\, R^{\mathrm{base}}_i + \lambda_c\, s^{c}_i\big),
\qquad \lambda_a{=}0.7,\ \lambda_c{=}0.3,
\label{eq:app-master}
\end{equation}
where $R^{\mathrm{base}}_i\in[0,1]$ is task accuracy plus a lightweight format check (the task term $R_{\text{base}}$), and $s^{c}_i\in[0,1]$ is the unified decision reward defined below. We write $G_{\text{exp}}$ and $G_{\text{direct}}$ for the exploring and direct-answering subsets of $\mathcal{G}$. Both rubric judges, i.e., the query-quality judge of $R_{\text{search}}$ and the grounding judge of $R_{\text{skill}}$, are Qwen3.7-Max. All LLM judgments are zero-shot calls returning JSON, from which we parse integer ratings and clip every score to $[0,1]$.

\subsection{When to Search ($R_{\text{when}}$): Group-Contrastive Gate}\label{app:reward-when}

We judge exploration necessity \emph{relatively} within each group rather than with a fixed bonus. Let $\bar{y}_{\text{exp}}$ and $\bar{y}_{\text{direct}}$ be the mean accuracies of $G_{\text{exp}}$ and $G_{\text{direct}}$, and $\Delta=\bar{y}_{\text{exp}}-\bar{y}_{\text{direct}}$ their gain. With margin $\tau_0{=}0.25$,
\begin{equation}
s^{c}_i =
\begin{cases}
q_i, & \Delta > \tau_0 \ \text{ and } \ i \in G_{\text{exp}}, \\[2pt]
1, & \Delta < -\tau_0 \ \text{ and } \ i \in G_{\text{direct}}, \\[2pt]
0, & \text{otherwise}.
\end{cases}
\label{eq:app-contrastive}
\end{equation}
The key property we enforce is that the quality magnitude $q_i$ is granted \emph{only} to exploring rollouts, and \emph{only} when exploration clearly beats direct answering ($\Delta>\tau_0$); if exploration does not outperform direct answering, it earns no quality bonus at all. This ties the quality reward to a demonstrated capability gap and calibrates exploration rather than encouraging it indiscriminately. (We use subset means whenever both strategies appear in the group; if either subset is empty, $s^{c}_i{=}0$.)

\subsection{Quality Magnitude $q_i$ ($R_{\text{search}}$ \& $R_{\text{skill}}$)}\label{app:reward-quality}

The magnitude $q_i$ folds skill quality and search discipline into a single expression:
\begin{equation}
q_i \;=\; \max\!\Big(0,\; \tfrac{1}{2}\big(s^{\mathrm{reuse}}_i + s^{\mathrm{ground}}_i\big) - p^{\mathrm{q}}_i\Big),
\label{eq:app-quality}
\end{equation}
where each skill signal lies in $[0,1]$ and the query penalty in $[0,0.3]$. Recall from Eq.~\ref{eq:app-contrastive} that $q_i$ takes effect only for exploring rollouts that beat direct answering, so a bonus is paid only for a skill that is reusable, grounded, and reached through well-posed queries. The three components are:

\begin{itemize}[leftmargin=1.2em, itemsep=2pt]
    \item \textbf{Skill Reuse ($s^{\mathrm{reuse}}_i$).} Skill generalizability: the skill is injected into $K{=}3$ similar held-out problems, each of which is solved twice by a frozen copy of the policy's own backbone---Qwen3-8B when training the 8B policy and Qwen3-4B when training the 4B policy---once with the generated skill available in its context and once by direct answering. We take the mean pass-rate difference between the two configurations as the gain $\bar{\delta}\in[-1,1]$, giving $s^{\mathrm{reuse}}_i=\operatorname{clip}_{[0,1]}(\tfrac{1}{2}+\tfrac{1}{2}\bar{\delta})$ (no change $\to 0.5$, helps $\to 1$, hurts $\to 0$).
    \item \textbf{Skill Grounding ($s^{\mathrm{ground}}_i$).} Source faithfulness: the judge performs a claim-level audit against the retrieved evidence and returns a $1$--$10$ score, which we normalize to $[0,1]$.
    \item \textbf{Query Penalty ($p^{\mathrm{q}}_i$).} Search discipline: the judge scores each query set on two dimensions---\emph{query abstraction} (searching for principles, not specific numbers) and \emph{evidence gain} (returning concrete, non-redundant evidence); each dimension scoring $\le 1$ adds $0.15$, so $p^{\mathrm{q}}_i\in\{0,0.15,0.30\}$.
\end{itemize}

The two judge prompts follow.

\begin{judgebox}{Query-Quality Judge Prompt ($R_{\text{search}}$)}
You are a query-quality judge for an AI agent that answers questions using web search. The agent has generated a set of Google search queries during exploration. Your job is to evaluate whether these queries retrieve useful evidence to solve the question \emph{from the agent's current uncertainty}. Do NOT judge whether the final answer is correct; do NOT use the ground-truth answer.

\smallskip
\textbf{Inputs.}
\begin{itemize}[leftmargin=1.2em,itemsep=1pt,topsep=1pt]
    \item \textbf{Question:} \texttt{\{question\}}
    \item \textbf{Candidate options (if any):} \texttt{\{options\}}
    \item \textbf{Agent goal:} \texttt{\{exploration\_goal\}}
    \item \textbf{Agent reasoning before query:} \texttt{\{agent\_reasoning\_before\_query\}}
    \item \textbf{Queries to judge:} \texttt{\{queries\_json\}}
    \item \textbf{Search snippets returned:} \texttt{\{search\_snippets\}}
\end{itemize}

\smallskip
\textbf{Scoring dimensions.} Score the query evaluation on the following $2$ dimensions. Each dimension must be an integer score: $0$, $1$, $2$, or $3$.

\smallskip
\textbf{1. Query Abstraction.} Whether the query searches for reusable principles, formulas, or workflows, rather than over-fitting to specific instance numbers or searching vague broad topics.
\begin{itemize}[leftmargin=1.2em,itemsep=1pt,topsep=1pt]
    \item $0$ = \emph{Instance-Overfitted}: query contains specific problem numbers or local variables (e.g.\ ``calculate X when a=3, b=5''), directly fishing for the answer.
    \item $1$ = \emph{Broad Topic}: query is too vague, using only top-level domain keywords (e.g.\ ``physics thermodynamics'').
    \item $2$ = \emph{Concrete Target}: targets the right uncertainty, but still carries specific numbers or instance context from the problem.
    \item $3$ = \emph{De-instanced Precision}: strips away problem-specific numbers/entities and uses precise domain terminology to search for reusable formulas or standard methods.
\end{itemize}

\smallskip
\textbf{2. Evidence Gain.} Whether the returned search snippets eliminate confusion and provide verifiable, non-redundant evidence.
\begin{itemize}[leftmargin=1.2em,itemsep=1pt,topsep=1pt]
    \item $0$ = \emph{Noisy/Irrelevant}: snippets are empty, noisy, or completely useless.
    \item $1$ = \emph{Redundant Background}: snippets only give common background knowledge the agent already knew.
    \item $2$ = \emph{Partial Support}: snippets provide useful formulas or rules that narrow down the space.
    \item $3$ = \emph{Crucial Gain}: snippets provide direct conceptual proof or clear counter-evidence to resolve the core uncertainty.
\end{itemize}

\smallskip
\textbf{Output (return ONLY valid JSON).}
\begin{lstlisting}
{
  "dimension_scores": {
    "query_abstraction": <integer 0-3>,
    "evidence_gain": <integer 0-3>
  },
  "rationale": "1-3 sentences explaining the main strengths or weaknesses."
}
\end{lstlisting}
\end{judgebox}

\begin{judgebox}{Skill Grounding Judge Prompt ($R_{\text{skill}}$)}
You are a strict skill-quality reviewer for a research project that uses LLM agents to ``explore'' the web and then distill a reusable skill (workflow + python helper functions). You will rate ONE skill on a SINGLE dimension---Grounding---on an integer scale $1$--$10$.

\smallskip
\textbf{Dimension --- Grounding (source faithfulness).} How much of the skill (workflow + code) is genuinely supported by the evidence the rollout retrieved during exploration?
\begin{itemize}[leftmargin=1.2em,itemsep=1pt,topsep=1pt]
    \item \textbf{9--10} = every important claim (formula, fact, constant, procedure, edge case) is supported by the evidence; no fabrication.
    \item \textbf{7--8} = most important claims are supported; a few minor items rely on prior knowledge.
    \item \textbf{5--6} = roughly half the skill is supported by evidence; the rest is prior knowledge or unverifiable.
    \item \textbf{3--4} = mostly prior knowledge dressed up as if evidence-derived; the evidence is barely used.
    \item \textbf{1--2} = skill fabricates content, misreads the evidence, or copies wrong numbers/units from the evidence.
\end{itemize}

\smallskip
\textbf{Audit instructions.}
\begin{itemize}[leftmargin=1.2em,itemsep=1pt,topsep=1pt]
    \item Pick the most important ``core claims'' the skill makes (formulas, numeric constants/conversion factors, named procedures, edge-case rules). Aim for $4$--$8$ claims; skip pure boilerplate.
    \item For each, decide: \texttt{supported} (clearly supported somewhere in the retrieved evidence; quote a SHORT snippet $\le 25$ words); \texttt{prior} (not in the evidence at all; the model is using its own prior knowledge); \texttt{fabricated} (contradicts the evidence, misreads it, or copies a value with the wrong unit/dimension; briefly say how).
    \item Be specific about numbers/units.
\end{itemize}

\smallskip
\textbf{Output (STRICT JSON, NO prose outside).}
\begin{lstlisting}
{
  "score": <1-10>,
  "reason": "<= 120 words; explain the score using the audit findings",
  "claims": [
    {"claim": "<paraphrased <= 25 words>",
     "verdict": "supported" | "prior" | "fabricated",
     "evidence_snippet": "<short quote <= 25 words; empty if not supported>",
     "note": "<= 25 words; optional clarification>"}
  ]
}
\end{lstlisting}
\end{judgebox}

\paragraph{Reward variants and constants.}
The reward ablations in the main paper remove one decision reward at a time while keeping the rest of the pipeline fixed: \emph{w/o} $r_{\mathrm{when}}$ drops the group-contrastive gate, \emph{w/o} $r_{\mathrm{sea}}$ drops the query penalty $p^{\mathrm{q}}_i$, \emph{w/o} $r_{\mathrm{gen}}$ drops the skill-quality terms $s^{\mathrm{reuse}}_i$ and $s^{\mathrm{ground}}_i$, and \emph{w/o all} keeps only the task term $R_{\text{base}}$. All constants are held fixed across runs: $N{=}16$, $\lambda_a{=}0.7$, $\lambda_c{=}0.3$, $\tau_0{=}0.25$, $K{=}3$, and a per-dimension query penalty of $0.15$.

\section{Preliminary Study Implementation Details}\label{app:preliminary}

\paragraph{Setup.}
We use 500 questions from SuperGPQA-Science with teacher-generated reference solutions; questions unsolved by the teacher are discarded.
The agents are Qwen3-8B/14B/32B/235B-A22B, and Qwen3.7-Max is used as the teacher judge throughout.

\paragraph{Pipeline.}
Each question is processed in four steps:
\begin{enumerate}[leftmargin=1.4em, itemsep=2pt]
    \item \textbf{Self-assessment.} In one context, the agent decides whether to trigger exploration and, if yes, states the capability or knowledge it believes is missing.
    \item \textbf{Direct answering.} In a separate fresh context, the agent answers the question with standard chain-of-thought prompting, without seeing its self-assessment output. We sample several attempts per question and count the question as failed only when all of them are wrong.
    \item \textbf{Teacher diagnosis.} For a failed question, all sampled attempts are passed to the teacher in a single call, and the teacher compares them against the reference solution to identify the recurring primary cause of failure.
    \item \textbf{Upper-bound crediting.} For incorrectly answered questions that triggered exploration, the teacher judges whether exploring the self-assessed gap would plausibly have covered the diagnosed cause of failure; only such questions are credited toward the upper bound.
\end{enumerate}

We report two quantities from this pipeline. First, the \emph{exploration trigger rate}, computed separately over the questions the agent eventually answers correctly and incorrectly in Step~2, which measures how well the agent anticipates its own failures. Second, the \emph{execution accuracy}, i.e., the direct-answering accuracy of Step~2, together with the \emph{upper bound} of self-evolution, which additionally credits the questions that are answered incorrectly but trigger exploration and receive a \texttt{MATCH} in Step~4.

The prompts used in Steps~1, 3, and 4 are given below.

\begin{tcolorbox}[enhanced jigsaw, breakable, colback=white, colframe=black!70, colbacktitle=black!10, coltitle=black, fonttitle=\bfseries, fontupper=\footnotesize, arc=6pt, title={Step 1: Self-Assessment Prompt (Student)}]
You are an intelligent assistant solving a problem.

\smallskip
\textbf{Question:} \texttt{\{question\}}

\smallskip
\textbf{Rules.}
\begin{itemize}[leftmargin=1.2em,itemsep=1pt,topsep=1pt]
    \item Think step-by-step, then output exactly one action tag.
    \item Do not guess when uncertain ($<$90\% confidence $\Rightarrow$ explore).
    \item Default to explore for problems involving theorems, specialized techniques, or domain-specific formulas.
\end{itemize}

\smallskip
\textbf{Uncertainty signals (any $\Rightarrow$ must explore):} unsure about specific facts/formulas; torn between options; relying on vague memory; problem requires a method you cannot confidently execute.

\smallskip
\textbf{Actions (choose one, only \texttt{<ANSWER>} when confident).}
\begin{lstlisting}
<EXPLORE>specific knowledge gap</EXPLORE>
<ANSWER>final answer</ANSWER>
\end{lstlisting}
\end{tcolorbox}

\begin{tcolorbox}[enhanced jigsaw, breakable, colback=white, colframe=black!70, colbacktitle=black!10, coltitle=black, fonttitle=\bfseries, fontupper=\footnotesize, arc=6pt, title={Step 3: Teacher Diagnosis Prompt}]
A student model attempted the following problem several times but got it wrong every time.

\smallskip
\textbf{Problem:} \texttt{\{question\}} \quad 

\textbf{Ground Truth:} \texttt{\{ground\_truth\}}\\
\textbf{Correct Solution:} \texttt{\{solution\}} \quad

\textbf{Student Attempts (all sampled trajectories):} \texttt{\{multi\_student\_answers\}}

\smallskip
Identify the single most critical knowledge gap or reasoning error that recurs across the attempts and caused the failure. Be specific---name the exact theorem, concept, or technique.

\smallskip
\textbf{Output:}
\begin{lstlisting}
<CORE_DIAGNOSIS>one sentence</CORE_DIAGNOSIS>
\end{lstlisting}
\end{tcolorbox}

\begin{tcolorbox}[enhanced jigsaw, breakable, colback=white, colframe=black!70, colbacktitle=black!10, coltitle=black, fonttitle=\bfseries, fontupper=\footnotesize, arc=6pt, title={Step 4: Upper-Bound Crediting Prompt (Teacher)}]
\textbf{Problem:} \texttt{\{question\}}\\
\textbf{Model's Self-Assessed Gap (before attempting):} \texttt{\{self\_gap\}}\\
\textbf{Teacher's Diagnosed Cause of Failure (after failure):} \texttt{\{teacher\_gap\}}

\smallskip
Judge whether exploring the self-assessed gap would have plausibly covered or prevented the diagnosed failure. The self-assessment may be broader or use different words---that's fine.

\smallskip
\textbf{Output:}
\begin{lstlisting}
<MATCH>YES</MATCH> or <MATCH>NO</MATCH>
\end{lstlisting}
\end{tcolorbox}

\section{Failure Pattern Analysis Details}\label{app:error-analysis}

This section complements the error analysis in the main paper (RQ1) by giving the full definition of each of the five failure patterns; for three of them we additionally show a representative case drawn from the evaluation trajectories. The patterns are assigned by Qwen3.7-Max, which inspects each trajectory and labels the failure it exhibits.

\paragraph{(1) Missed Search Trigger.}
The agent answers directly on a question that exceeds its parametric knowledge, relying on vaguely remembered facts instead of triggering exploration. Since skipping search can occur on any question, this pattern is measured over all questions; the remaining four are conditioned on questions where exploration is triggered.

\paragraph{(2) Poor Search Query.}
Exploration is triggered, but the queries miss the true information need: they pursue a wrong hypothesis (confirmation bias), are too generic to retrieve new evidence, or omit the key terminology that separates the candidate options.
Figure~\ref{fig:case_poor_search_query_gm_counter} shows a representative case where the queries miss the required definition.

\begin{figure}[htbp]
\centering
\small
\begin{tcolorbox}[colback=gray!5, colframe=gray!60, arc=2mm, boxrule=0.8pt, title=\textbf{Poor Search Query: Missing the Definition}, fonttitle=\small\sffamily, left=4pt, right=4pt, top=4pt, bottom=4pt]
\small

\textbf{Question.}
\noindent The plateau slope of a Geiger--M\"uller counter tube is 5\%, with a plateau length of 250 V. When the operating voltage is 175 V above the start point rather than 50 V above it, by how much does the count rate increase?

\vspace{5pt}
\hrule
\vspace{5pt}

\textbf{Model Search Queries.}
\noindent\textit{``Geiger-M\"uller counter plateau slope relationship with count rate''}

\noindent\textit{``how does voltage affect count rate in GM tube plateau region''}

\vspace{5pt}
\hrule
\vspace{5pt}

\textbf{Why the Query Is Poor.}
\noindent The question is not asking for a qualitative relationship between voltage and count rate. It requires the precise definition of \textbf{plateau slope}, i.e., how a ``5\% slope'' is quantified in a GM counter plateau calculation.

\vspace{5pt}
\hrule
\vspace{5pt}

\textbf{Better Search Query.}
\noindent\textit{``GM counter plateau slope definition percent per 100V''}

\noindent\textit{``Geiger Muller plateau slope formula count rate per 100 volts''}

\end{tcolorbox}
\caption{\textbf{Poor search query.} The model searches for a broad qualitative relationship, while the task requires the operational definition of ``plateau slope'' used in GM-counter calculations.}
\label{fig:case_poor_search_query_gm_counter}
\end{figure}

\paragraph{(3) Formatting Error.}
The agent violates the action protocol, e.g., looping over \texttt{<QUERY>} without visiting pages or generating a skill, emitting malformed action tags, or exhausting the round budget, so the episode ends without a usable outcome.

\paragraph{(4) Insufficient Skill Coverage.}
The skill library fails to supply the knowledge the question requires: no relevant skill exists or is retrieved, the retrieved skill targets a mismatched scope, or the skill is written too specifically to its source question to generalize to the new one.
Note that Search2Skill does not train the retrieval component itself; the reduction of this failure category comes from calibrated skill \emph{generation}: RL drives the model to produce more high-quality skills with accurate applicability descriptions (\texttt{use\_when}) on questions it fails, which enriches the coverage of the skill library and makes the fixed retriever more likely to surface a usable skill.
Figure~\ref{fig:case_skill_retrieval_attempted_murder} illustrates such a generalization failure.

\begin{figure}[htbp]
\centering
\small
\begin{tcolorbox}[colback=gray!5, colframe=gray!60, arc=2mm, boxrule=0.8pt, title=\textbf{Insufficient Skill Coverage: Scope-Mismatched Skill}, fonttitle=\small\sffamily, left=4pt, right=4pt, top=4pt, bottom=4pt]
\small

\textbf{Question.}
\noindent A defendant drives over 100 MPH through a residential neighborhood, loses control, and severely injures a pedestrian. If charged with attempted murder, is he guilty?

\vspace{5pt}
\hrule
\vspace{5pt}

\textbf{Retrieved Skill.}
\noindent The system retrieves \texttt{murder\_intent\_recklessness\_analysis}, whose workflow states that murder may require intent to kill \textit{or extreme recklessness}, emphasizing deadly-weapon and recklessness reasoning.

\vspace{5pt}
\hrule
\vspace{5pt}

\textbf{Coverage Gap.}
\noindent The retrieved skill is valid for completed murder, but the library contains no skill encoding the narrower rule that \textbf{attempted murder requires specific intent to kill}. The agent therefore transfers a recklessness rule to an attempt offense where it does not apply, and wrongly convicts the defendant.

\end{tcolorbox}
\caption{\textbf{Insufficient skill coverage.} A semantically related homicide skill is retrieved, but its scope does not cover the specific-intent requirement of attempted murder.}
\label{fig:case_skill_retrieval_attempted_murder}
\end{figure}

\paragraph{(5) Skill Hallucination.}
The generated skill is unfaithful to the retrieved evidence---it fabricates rules, distorts definitions, or encodes a wrong conclusion as a reusable fact. The agent then follows its own corrupted skill to a wrong answer, and the polluted skill can mislead future questions that retrieve it.
Figure~\ref{fig:case_skill_hallucination_gap_loss} shows a case where useful retrieved evidence is distilled into an unsupported formula.

\begin{figure}[htbp]
\centering
\small
\begin{tcolorbox}[colback=gray!5, colframe=gray!60, arc=2mm, boxrule=0.8pt, title=\textbf{Skill Hallucination: Correct Clue, Fabricated Formula}, fonttitle=\small\sffamily, left=4pt, right=4pt, top=4pt, bottom=4pt]
\small

\textbf{Question.}
\noindent A playback head has gap width $1.0\times10^{-5}$ m; the recorded wavelength is $2.0\times10^{-5}$ m. Determine the gap loss.

\vspace{5pt}
\hrule
\vspace{5pt}

\textbf{Retrieved Evidence.}
\noindent The retrieved pages correctly attributed gap loss to the \textbf{finite gap length} of the reproduce head and pointed to the standard \textit{$\sin x/x$}-type transfer-function formula.

\vspace{5pt}
\hrule
\vspace{5pt}

\textbf{Generated Skill (Hallucinated).}
\noindent The skill instead invented an unsupported logarithmic rule
$\text{Gap Loss} = 20\log_{10}\!\big(\lambda/2g\big)$,
which is found nowhere in the evidence. Plugging in the given values yields $0$ dB---matching no option---so the model drifted to a nearby simple-ratio answer ($6.1$ dB) instead of the correct $3.9$ dB.

\end{tcolorbox}
\caption{\textbf{Skill hallucination.} Useful retrieved evidence about finite-gap loss is converted into an unsupported simplified logarithmic formula, and the model solves the numerical problem with the wrong rule.}
\label{fig:case_skill_hallucination_gap_loss}
\end{figure}

\begin{table*}[t]
\centering
\small
\setlength{\tabcolsep}{4pt}
\begin{tabular*}{\textwidth}{@{\extracolsep{\fill}}l ccc ccc c@{}}
\toprule
\multirow{2}{*}{\textbf{Method}}
  & \multicolumn{3}{c}{\textbf{SuperGPQA}}
  & \multicolumn{3}{c}{\textbf{MMLU-Pro}}
  & \multirow{2}{*}{\textbf{Avg}} \\
\cmidrule(lr){2-4}\cmidrule(lr){5-7}
  & Mathematics & Management & Science & Law & Philosophy & History & \\
\midrule
Direct Inference & 64.8\textsubscript{$\pm$0.75} & 41.4\textsubscript{$\pm$1.33} & 45.6\textsubscript{$\pm$1.60} & 39.5\textsubscript{$\pm$0.54} & 58.3\textsubscript{$\pm$1.20} & 58.8\textsubscript{$\pm$1.06} & 51.4\textsubscript{$\pm$0.61} \\
Search Agent & 67.2\textsubscript{$\pm$1.51} & 43.0\textsubscript{$\pm$1.15} & 50.2\textsubscript{$\pm$0.12} & 35.8\textsubscript{$\pm$1.21} & 60.9\textsubscript{$\pm$1.00} & 59.6\textsubscript{$\pm$0.95} & 52.8\textsubscript{$\pm$0.72} \\
Search Agent\textsubscript{train} & 67.0\textsubscript{$\pm$0.81} & 44.0\textsubscript{$\pm$2.75} & 62.0\textsubscript{$\pm$0.95} & 43.6\textsubscript{$\pm$0.96} & 68.3\textsubscript{$\pm$1.28} & 60.6\textsubscript{$\pm$0.80} & 57.7\textsubscript{$\pm$0.77} \\
ReasoningBank & 64.6\textsubscript{$\pm$1.67} & 38.0\textsubscript{$\pm$1.22} & 48.8\textsubscript{$\pm$2.83} & 25.0\textsubscript{$\pm$1.42} & 55.9\textsubscript{$\pm$1.64} & 57.7\textsubscript{$\pm$0.53} & 48.3\textsubscript{$\pm$1.05} \\
Memp & 67.2\textsubscript{$\pm$0.40} & 41.0\textsubscript{$\pm$0.99} & 45.6\textsubscript{$\pm$3.41} & 27.7\textsubscript{$\pm$3.21} & 60.9\textsubscript{$\pm$0.61} & 57.5\textsubscript{$\pm$1.06} & 49.9\textsubscript{$\pm$0.56} \\
EvolveR\textsubscript{train} & 64.0\textsubscript{$\pm$1.07} & 46.0\textsubscript{$\pm$2.16} & 63.8\textsubscript{$\pm$1.30} & 41.2\textsubscript{$\pm$0.86} & 67.5\textsubscript{$\pm$0.82} & 61.9\textsubscript{$\pm$0.73} & 57.4\textsubscript{$\pm$0.64} \\
Search2Skill & 68.0\textsubscript{$\pm$0.92} & 46.0\textsubscript{$\pm$1.31} & 48.0\textsubscript{$\pm$1.11} & 38.4\textsubscript{$\pm$0.97} & 62.3\textsubscript{$\pm$1.91} & 59.6\textsubscript{$\pm$0.66} & 53.7\textsubscript{$\pm$0.38} \\
\textbf{Search2Skill\textsubscript{train}} & \textbf{71.6}\textsubscript{$\pm$0.53} & \textbf{50.0}\textsubscript{$\pm$1.56} & \textbf{63.2}\textsubscript{$\pm$0.23} & \textbf{45.4}\textsubscript{$\pm$2.76} & \textbf{68.7}\textsubscript{$\pm$1.14} & \textbf{65.1}\textsubscript{$\pm$1.22} & \textbf{60.7}\textsubscript{$\pm$0.83} \\
\bottomrule
\end{tabular*}
\caption{Streaming results on Qwen3-8B reported as mean\,$\pm$\,standard deviation over three independent runs (question order and sampling seed $42$, $67$, $92$).}
\label{tab:app-variance}
\end{table*}

\section{Run-to-Run Variance}\label{app:variance}

Every streaming number in the main paper is the mean over three independent runs, each using a different random question order together with a different sampling seed (seeds $42$, $67$, $92$; decoding temperature $0.3$). Because decoding is stochastic, even order-independent methods such as Direct Inference and Search Agent fluctuate across runs. Table~\ref{tab:app-variance} repeats the Qwen3-8B streaming results with the standard deviation over these three runs. EvoAgentBench is evaluated only under the held-out protocol and therefore does not appear here.

\section{Prompt Templates}\label{app:prompts}

We list the three prompt templates that drive Search2Skill's agent loop: (i)~the decision prompt (Prompt~1), under which the agent chooses among activating a skill, running code, exploring externally, or answering directly; (ii)~the exploration prompt (Prompt~2), invoked once the agent decides to explore, which structures the web-search, page-visit, and skill-generation actions; and (iii)~the skill-only execution prompt (Prompt~3), used at evaluation time to test the reusability of skills accumulated previously without further exploration. Placeholders in braces (e.g.\ \texttt{\{question\}}, \texttt{\{formatted\_skills\}}) are filled at runtime.

\subsection{Decision Prompt}\label{app:prompt-decision}

Prompt~1 is the entry point of every problem-solving episode. It enumerates the uncertainty signals that should trigger external exploration and exposes the four action tags the agent can emit. The exploration-necessity reward in our RL scheme is computed from the action tag chosen under exactly this prompt.

\begin{promptbox}
You are an intelligent assistant solving problems in multiple rounds.

\smallskip
\textbf{Available Skills:} \texttt{\{formatted\_skills\}}

\smallskip
\textbf{Question:} \texttt{\{question\}}

\smallskip
\textbf{Rules}
\begin{itemize}[leftmargin=1.2em,itemsep=1pt,topsep=1pt]
    \item Think step-by-step in plain text, then output exactly one action tag at the end.
    \item You will receive feedback after each action.
    \item Never output raw code directly; always use \texttt{<CODE>} to run Python code.
    \item \textbf{Critical:} do not guess when uncertain. If you are not highly confident ($>$90\%) in your answer, you must use \texttt{<EXPLORE>} to search for evidence first. Getting the answer wrong is worse than spending an extra round exploring.
\end{itemize}

\smallskip
\textbf{Uncertainty signals (any of these means you must explore).}
\begin{itemize}[leftmargin=1.2em,itemsep=1pt,topsep=1pt]
    \item The question asks about specific facts, names, dates, numbers, or formulas you are not 100\% sure about.
    \item You do not know or are not fully sure about a concept, knowledge point, procedure, workflow, API, library usage, command, syntax, or implementation detail needed to solve the task.
    \item You find yourself torn between two or more options.
    \item You are relying on vague memory rather than solid knowledge.
    \item The problem requires a specific algorithm, technique, or mathematical method that you are not fully confident you can implement correctly (e.g., a DP variant, number-theoretic formula, geometric algorithm, or specialized data structure).
\end{itemize}

\smallskip
\textbf{Actions (choose exactly one).}
\begin{enumerate}[leftmargin=1.5em,itemsep=1pt,topsep=1pt]
    \item Activate a skill: \texttt{<ACTIVATE\_SKILL>}\allowbreak\texttt{skill\_name}\allowbreak\texttt{</ACTIVATE\_SKILL>}. Once you have activated a skill, you can view the knowledge contained within it, or directly utilize the functions included in the skill within \texttt{<CODE>}.
    \item Run Python code: \texttt{<CODE>your python code here</CODE>}.
    \item Explore~--~use this when uncertain or missing knowledge: \texttt{<EXPLORE>what specific information, knowledge, method, example, workflow, or implementation detail you need to verify</EXPLORE>}.
    \item Answer~--~only when confident: \texttt{<ANSWER>final answer (or option letter like B for multiple choice)</ANSWER>}.
\end{enumerate}
\end{promptbox}

\subsection{Exploration Prompt}\label{app:prompt-explore}

Once the agent emits an \texttt{<EXPLORE>} tag in Prompt~1, control transfers to Prompt~2, which scopes the exploration phase. This prompt enforces the search--visit--distill discipline that underlies Search2Skill and produces the JSON skill object against which the skill-quality reward is evaluated.

\begin{promptbox}
You are in the exploration phase. Goal: gather evidence to answer the question. Do not restate the question. Previous context is available.

\smallskip
\textbf{Current goal:} \texttt{\{exploration\_goal\}}

\smallskip
\textbf{Rules.}
\begin{itemize}[leftmargin=1.2em,itemsep=1pt,topsep=1pt]
    \item Think step-by-step in plain text, then output exactly one action tag at the end.
    \item You must generate a skill before answering or calling other skills.
\end{itemize}

\smallskip
\textbf{Search rules~--~what to search for.}
\begin{itemize}[leftmargin=1.2em,itemsep=1pt,topsep=1pt]
    \item Never search for the problem itself or copy-paste the question into queries.
    \item Instead, search for the underlying knowledge needed to solve it: formulas, theorems, or domain knowledge (e.g.\ ``VSEPR theory bond angles''); library/API usage (e.g.\ ``python selfies library decode example'', ``rdkit MolFromSmiles tutorial''); algorithms, techniques, or mathematical methods (e.g.\ ``digit DP tutorial with modulo constraint'', ``monotonic stack range query algorithm'', ``Euler totient modular inverse'', ``how to convert SELFIES to SMILES python'').
    \item Your goal is to learn the method or tool, then apply it yourself.
\end{itemize}

\smallskip
\textbf{Recommended exploration chain (follow this order).}
\textbf{Step~1.} \texttt{<QUERY>}~--~search for methods, formulas, or library/API usage relevant to the problem type. 

\textbf{Step~2.} \texttt{<PAGE\_VISIT>}~--~pick the most relevant URLs from search results and visit them to read full content with \texttt{[Supplementary knowledge]}. Search snippets are too short to be reliable; you must visit pages to get accurate, detailed information before generating a skill. 

\textbf{Step~3.} \texttt{<GENERATE\_SKILL>}~--~after multiple rounds of \texttt{<QUERY>} and \texttt{<PAGE\_VISIT>}, once you believe you have gathered sufficient information, you can synthesize the detailed page content into a reusable skill.

\smallskip
\textbf{Critical:} do not skip \texttt{<PAGE\_VISIT>}. Search snippets are brief summaries that often lack critical details or contain misleading fragments. Always visit at least one high-quality page to read the full content before generating a skill.

\smallskip
\textbf{Actions (choose exactly one per round).}
\begin{enumerate}[leftmargin=1.5em,itemsep=1pt,topsep=1pt]
    \item Search web (at most 3 queries per call): \texttt{<QUERY>["query 1", "query 2"]</QUERY>}.
    \item Visit pages to read full content (at most 4 URLs per call): \texttt{<PAGE\_VISIT>\{"urls": ["url1", "url2"], "goal": "what to extract"\}</PAGE\_VISIT>}.
    \item Generate skill (only after visiting pages):
\begin{lstlisting}
<GENERATE_SKILL>{
  "skill_name": "name_with_underscores",
  "use_when": "when to activate this skill -- describe the problem type or scenario",
  "workflow": "step-by-step workflow, distilled facts, formulas, constants from page content (or empty string)",
  "code": "reusable Python `def` functions (or empty string)"
}</GENERATE_SKILL>
\end{lstlisting}

\end{enumerate}

\smallskip
\textbf{Skill design rules (\texttt{<GENERATE\_SKILL>}).}
\begin{itemize}[leftmargin=1.2em,itemsep=1pt,topsep=1pt]
    \item Must contain \texttt{skill\_name} and at least one non-empty \texttt{workflow} or \texttt{code}.
    \item \textbf{Generalization rule.} Do not just solve the immediate question. Synthesize all uncovered domain knowledge into a broadly applicable skill for similar future problems. \textbf{Critical:} you must extract non-obvious domain insights from the \texttt{[Supplementary knowledge]} sections in page-visit results and include them.
    \item \texttt{workflow} (facts \& insights): distilled facts, formulas, rules, and step-by-step procedures.
    \item \texttt{code} (computation \& logic): only \texttt{def} functions (no top-level execution/prints). Loaded as a library. Use for complex math or external libs. Encourage parameterized functions: every function must accept inputs as arguments and compute results dynamically. Forbidden returning a hardcoded constant; never write a function whose sole purpose is to return a fixed value (e.g., \texttt{def get\_gravity(): return 9.8}). Such facts belong in \texttt{workflow}, not in code.
    \item \texttt{code} functions must be self-contained and not rely on any external fetching helpers. All reference data should be distilled into \texttt{workflow} instead.
    \item \textbf{Groundedness rule.} All factual claims, formulas, and parameters in \texttt{workflow} and \texttt{code} should be based on content you read via \texttt{<PAGE\_VISIT>}. Do not substitute your internal knowledge for domain-specific facts. If a fact was not found in the visited pages, omit it rather than guess.
\end{itemize}
\smallskip
\textbf{Example~--~a skill combining workflow and computation.}
(Line breaks within \texttt{"workflow"} and \texttt{"code"} are rendered visually for readability; the underlying JSON string uses \texttt{\textbackslash n}.)

\begin{lstlisting}
<GENERATE_SKILL>{
  "skill_name": "radionuclide_decay_solver",
  "use_when": "Solving radioactive decay problems: look up half-lives and compute remaining quantities",
  "workflow": "Decay formula: N = N0 * (1/2)^(t / t_half).
Common half-lives:
  C-14: 5730 years, U-238: 4.468e9 years,
  K-40: 1.248e9 years, Ra-226: 1600 years,
  Co-60: 5.27 years, I-131: 8.02 days,
  Sr-90: 28.8 years.",
  "code": "def decay(n0, half_life, t):
    return n0 * (0.5 ** (t / half_life))

def time_to_fraction(fraction, half_life):
    import math
    return half_life * math.log2(1 / fraction)"
}</GENERATE_SKILL>
\end{lstlisting}

\end{promptbox}

\subsection{Skill-Only Execution Prompt}\label{app:prompt-internal}

Prompt~3 is used in the skill-transferability experiments (Held-out setting) and in any evaluation that disables further exploration. It removes the \texttt{<EXPLORE>} action and grounds the agent's reasoning entirely on the skills accumulated during training.

\begin{promptbox}
You are an intelligent assistant solving problems using skills. Exploration is disabled.

\smallskip
\textbf{Available Skills:} \texttt{\{formatted\_skills\}}

\smallskip
\textbf{Question:} \texttt{\{question\}}

\smallskip
\textbf{Rules.}
\begin{itemize}[leftmargin=1.2em,itemsep=1pt,topsep=1pt]
    \item Think step-by-step in plain text, then output exactly one action tag at the end.
    \item You will receive feedback after each action.
    \item Never output raw code directly; always use \texttt{<CODE>} to run Python code.
\end{itemize}

\smallskip
\textbf{Actions (choose exactly one).}
\begin{enumerate}[leftmargin=1.5em,itemsep=1pt,topsep=1pt]
    \item Activate a skill by outputting\\
    \texttt{<ACTIVATE\_SKILL>skill\_name</ACTIVATE\_SKILL>}.
    \item Run Python code with\\
    \texttt{<CODE>your python code here</CODE>}.
    \item Answer with\\
    \texttt{<ANSWER>final answer (or option letter like B for multiple choice)</ANSWER>}.
\end{enumerate}
\end{promptbox}

\section{Example Trajectory and Skill Examples}
\label{app:trajectory-skills}

\subsection{A Full Trajectory Example}
\label{app:full-trajectory-example}

We provide an illustrative example of the Search2Skill loop on a chemistry problem. In this trajectory, the agent identifies a missing piece of domain knowledge, enters the exploration phase to retrieve the relevant formula, distills the retrieved information into a reusable skill, and then invokes that skill to solve the original question.

\begin{trajbox}[title={Agent Trajectory: HCl Bond Length}]
\turnlabel{User}
The rotational constant of the $^{1}\mathrm{H}^{35}\mathrm{Cl}$ molecule, measured experimentally, is $B = 10.397~\mathrm{cm}^{-1}$. The reduced mass of this molecule is $1.63\times10^{-27}~\mathrm{kg}$. What is the equilibrium distance between the two atoms in the HCl molecule?

\smallskip
A.~0.140 nm \quad B.~0.135 nm \quad $\cdots$

\turnlabel{Assistant}
\begin{lstlisting}
<think>
I first recalled B = hc/(8 pi^2 I), but that can't be right for B in cm^-1 since the units don't match. I must verify the correct relation between B and I (and hence the bond length) before computing.
</think>

<EXPLORE>
What is the correct formula relating rotational constant B (in cm^-1) to moment of inertia I for a diatomic molecule? Also, what is the correct formula to compute equilibrium bond length R from B and reduced mass mu?
</EXPLORE>
\end{lstlisting}

\trajphase{Entering Exploration Mode}

\turnlabel{Assistant}
\begin{lstlisting}
<QUERY>
[
  "rotational constant B cm^-1
   moment of inertia diatomic",
  "bond length from rotational
   constant reduced mass formula"
]
</QUERY>
\end{lstlisting}

\turnlabel{Tool}
\begin{lstlisting}
Search results (top hits):
[1] cccbdb.nist.gov ...
[2] chem.libretexts.org ...
[3] ...
(snippets too brief -- visit pages)
\end{lstlisting}

\turnlabel{Assistant}
\begin{lstlisting}
<PAGE_VISIT>
{
 "urls": ["NIST CCCBDB",
          "LibreTexts", ...],
 "goal": "verify the B-I-R
          relation and units"
}
</PAGE_VISIT>
\end{lstlisting}

\turnlabel{Tool}
\begin{lstlisting}[style=toolbold]
Page visit results:

[1] cccbdb.nist.gov/convertmomint
The exact formula relating rotational
constant B (in cm^-1) to moment of
inertia I is
   @B = h / (8 pi^2 c I)@
where h is Planck's constant, c the
speed of light, and I the moment of
inertia.

I relates to bond length R and
reduced mass mu by
   @I = mu R^2@

Substituting I = mu R^2 gives
   B = h / (8 pi^2 c mu R^2)
therefore
   @R = sqrt(h / (8 pi^2 c mu B))@

Note: convert B from cm^-1 to m^-1
(x100) before using SI units.
\end{lstlisting}

\turnlabel{Assistant}
\begin{lstlisting}
<GENERATE_SKILL>
{
 "skill_name":
   "rot_constant_to_bond_length",
 "use_when":
   "Find bond length R from
    rotational constant B and
    reduced mass mu.",
 "workflow":
   "R = sqrt(h/(8 pi^2 c mu B));
    convert B: cm^-1 -> m^-1.",
 "code":
   "def bond_length_from_B(
       B_cm, mu): ..."
}
</GENERATE_SKILL>
\end{lstlisting}

\turnlabel{Tool}
\begin{lstlisting}
Skill registered:
  rot_constant_to_bond_length
Callable functions loaded.
\end{lstlisting}

\trajphase{Exiting Exploration Mode}

\turnlabel{Assistant}
\begin{lstlisting}[style=toolbold]
<think>
I now have the verified formula and
unit convention. I'll call the new
skill directly.
</think>

<CODE>
B_cm = 10.397     # cm^-1
mu   = 1.63e-27   # kg
R_m  = @bond_length_from_B(B_cm, mu)@
print(R_m * 1e9)  # nm
</CODE>
\end{lstlisting}

\turnlabel{Execution Result}
\begin{lstlisting}
0.12852125947517776
\end{lstlisting}

\turnlabel{Assistant}
\begin{lstlisting}
<think>
R ~ 0.1285 nm, matching option E
most closely.
</think>

<ANSWER>E</ANSWER>
\end{lstlisting}
\end{trajbox}

\subsection{Examples of Acquired Skills}
\label{app:skill-examples}

Below we show several additional acquired skills from different domains. Despite the diversity of task types, all skills follow the same standardized schema, with reusable knowledge captured in the \texttt{workflow} field and optional executable procedures captured in the \texttt{code} field. Depending on the domain, a skill may consist of executable helper functions, structured reasoning steps, or both.

\begin{skillbox}{Graph Theory}
\skillfield{skill\_name}\ \ \texttt{independent\_set\_counter}

\smallskip
\skillfield{use\_when}\ \ Counting independent sets of a given size in a grid graph or any graph, especially small cases.

\smallskip
\skillfield{workflow}\ \ An independent set is a set of vertices with no two adjacent. For a grid graph, vertices are adjacent if they share an edge (horizontal or vertical); diagonal adjacency is not considered. To count independent sets of size $k$, one simple approach is brute-force enumeration: generate all $k$-vertex combinations and check whether any selected pair is adjacent. For a $3\times3$ grid (9 vertices), there are $\binom{9}{3}=84$ possible 3-vertex combinations, so brute force is feasible. A more general alternative is backtracking: recursively add vertices while maintaining the independence constraint. For larger grids or general graphs, more advanced methods such as dynamic programming or transfer-matrix techniques may be useful, but brute-force enumeration is sufficient for small instances.

\smallskip
\skillfield{code}
\begin{lstlisting}
import itertools

def is_independent_set(graph, vertices):
    selected = set(vertices)
    for v in selected:
        for u in graph[v]:
            if u in selected:
                return False
    return True

def count_independent_sets_size_k(graph, k):
    vertices = list(graph.keys())
    count = 0
    for combo in itertools.combinations(vertices, k):
        if is_independent_set(graph, combo):
            count += 1
    return count

def generate_grid_graph(n):
    adj = {(i, j): set() for i in range(n) for j in range(n)}
    for i in range(n):
        for j in range(n):
            if i + 1 < n:
                adj[(i, j)].add((i + 1, j))
                adj[(i + 1, j)].add((i, j))
            if j + 1 < n:
                adj[(i, j)].add((i, j + 1))
                adj[(i, j + 1)].add((i, j))
    return adj
\end{lstlisting}
\end{skillbox}

\begin{skillbox}{Dynamic Programming}
\skillfield{skill\_name}\ \ \texttt{domino\_tiling\_3xn\_recurrence}

\smallskip
\skillfield{use\_when}\ \ Solving problems about counting domino tilings of a $3\times n$ rectangle with $2\times1$ or $1\times2$ dominoes.

\smallskip
\skillfield{workflow}\ \ For a $3\times n$ board tiled with dominoes, the number of tilings satisfies a linear recurrence. If $n$ is odd, the answer is zero because the board has area $3n$, which is odd and therefore cannot be fully covered by dominoes of area 2. For even $n$, two equivalent approaches are useful: a state-based dynamic program with auxiliary states, or the direct recurrence $f(n) = 4f(n-2) - f(n-4)$, with base cases $f(0)=1$, $f(2)=3$, and $f(4)=11$. For example, this recurrence gives $f(8)=153$. This skill is useful whenever a problem asks for the number of tilings of a $3\times n$ rectangle rather than a one-off small-case enumeration.

\smallskip
\skillfield{code}
\begin{lstlisting}
def count_3xn_domino_tilings(n):
    if n % 2 != 0:
        return 0
    if n == 0:
        return 1
    f = [0] * (n + 1)
    f[0] = 1
    if n >= 2:
        f[2] = 3
    if n >= 4:
        f[4] = 11
    for i in range(6, n + 1, 2):
        f[i] = 4 * f[i - 2] - f[i - 4]
    return f[n]

def count_3xn_domino_tilings_state(n):
    if n % 2 != 0:
        return 0
    if n == 0:
        return 1
    A = [0] * (n + 1)
    B = [0] * (n + 1)
    A[0] = 1
    B[0] = 0
    if n >= 1:
        A[1] = 0
        B[1] = 1
    for i in range(2, n + 1):
        A[i] = (A[i - 2] if i - 2 >= 0 else 0) + 2 * B[i - 1]
        B[i] = A[i - 1] + (B[i - 2] if i - 2 >= 0 else 0)
    return A[n]
\end{lstlisting}
\end{skillbox}

\begin{skillbox}{Biochemistry}
\skillfield{skill\_name}\ \ \texttt{enzyme\_optimal\_ph\_from\_pka}

\smallskip
\skillfield{use\_when}\ \ Determining the approximate optimal pH for an enzyme-catalyzed reaction given pKa values of ionizable groups in the active site and substrate, especially when activity depends on a specific protonation pattern.

\smallskip
\skillfield{workflow}\ \ The optimal pH of an enzyme-catalyzed reaction depends on the protonation states of key ionizable groups in the active site and substrate. In a simple two-group model where one group should be protonated and another should be deprotonated for maximal activity, the optimal pH is often approximated by the average of the two relevant pKa values. A more explicit approximation models activity as the product of the fraction of one group in the required protonation state and the fraction of the other group in its required state. This approximation is useful for reasoning about pH-activity profiles when a problem provides pKa values and asks for the likely pH optimum.

\smallskip
\skillfield{code}
\begin{lstlisting}
def optimal_ph_from_pka(pka_enzyme_group, pka_substrate):
    return (pka_enzyme_group + pka_substrate) / 2.0

def protonated_fraction(pH, pKa):
    return 1 / (1 + 10**(pH - pKa))

def deprotonated_fraction(pH, pKa):
    return 1 - protonated_fraction(pH, pKa)

def enzyme_activity(pH, pKa_histidine, pKa_substrate):
    histidine_protonated = protonated_fraction(pH, pKa_histidine)
    substrate_deprotonated = deprotonated_fraction(pH, pKa_substrate)
    return histidine_protonated * substrate_deprotonated
\end{lstlisting}
\end{skillbox}

\begin{skillbox}{Criminal Law}
\skillfield{skill\_name}\ \ \texttt{intoxication\_defenses\_criminal\_law}

\smallskip
\skillfield{use\_when}\ \ Determining which intoxication defense (voluntary, involuntary, insanity) is applicable and likely to succeed in criminal cases, especially murder charges.

\smallskip
\skillfield{workflow}
\begin{enumerate}[leftmargin=1.4em,itemsep=2pt,topsep=2pt]
    \item \textbf{Voluntary intoxication.} Defendant knowingly and intentionally consumes intoxicating substances. Only a \emph{partial} defense, applicable to specific-intent crimes (e.g.\ murder requiring premeditation/deliberation); not a defense to general-intent crimes. The defendant assumes the risk of impaired judgment.
    \item \textbf{Involuntary intoxication.} Defendant unknowingly or without consent ingests intoxicating substances (e.g.\ drugged without knowledge, tricked, coerced). A \emph{complete} defense to criminal liability, applicable to any charge. The defendant must prove by a preponderance of the evidence that the intoxication was not self-induced and negated the required criminal intent. It does not equate to insanity; it temporarily negates the mental state required for criminal responsibility.
    \item \textbf{Insanity defense.} Defendant suffers from a mental disease or defect that prevented forming specific intent or knowing/understanding the nature of the act. Requires proof by clear and convincing evidence that, at the time of the crime, the defendant could not appreciate the criminality of the conduct or conform conduct to the law. Drug-induced psychosis may qualify if it is a recognized mental disease and persists beyond acute intoxication (settled insanity).
    \item \textbf{Key distinctions.} Intoxication (voluntary/involuntary) is a temporary state affecting intent; insanity is a permanent or chronic mental disorder affecting rational thought. Involuntary intoxication can lead to a not-guilty-by-reason-of-insanity finding if the induced state meets insanity criteria, but the defense is typically framed as involuntary intoxication rather than insanity.
    \item \textbf{For murder.} Second-degree murder requires malice aforethought but not premeditation. Voluntary intoxication may negate specific intent for first-degree murder but not necessarily malice for second-degree. Involuntary intoxication is a complete defense, potentially leading to acquittal. The insanity defense may succeed if drug-induced psychosis is a recognized mental disease and the defendant lacked criminal capacity.
    \item \textbf{Burden of proof.} Defendant bears the burden for involuntary intoxication (preponderance); insanity carries a higher burden (clear and convincing).
\end{enumerate}

\smallskip
\skillfield{code}\ \ \textit{(none --- a knowledge-only skill).}
\end{skillbox}

\end{document}